\documentclass[twocolumn, switch]{article} 

\usepackage{preprint}

\usepackage{enumitem}
\usepackage{amsmath, amsthm, amssymb, amsfonts}
\usepackage{bm}
\usepackage{amsmath}
\usepackage{graphicx}
\usepackage{subfigure}
\usepackage{adjustbox}
\usepackage[algoruled,algo2e]{algorithm2e}

\usepackage{setspace}
\usepackage{amssymb}
\usepackage{booktabs}
\usepackage{array}
\usepackage{color}
\usepackage{multirow}
\usepackage{multicol}
\usepackage{threeparttable}
\usepackage{url}
\usepackage[page]{appendix}
\usepackage{makecell}

\usepackage[numbers,square]{natbib}
\usepackage[utf8]{inputenc}	
\usepackage[T1]{fontenc}	
\usepackage{xcolor}		
\usepackage[colorlinks = true,
            linkcolor = blue,
            urlcolor  = blue,
            citecolor = blue,
            anchorcolor = black]{hyperref}	
\usepackage{booktabs} 		
\usepackage{nicefrac}		
\usepackage{microtype}		
\usepackage{lineno}		
\usepackage{float}			

\usepackage{newfloat}
\usepackage{arydshln}
\DeclareFloatingEnvironment[name={Supplementary Figure}]{suppfigure}
\usepackage{sidecap}
\sidecaptionvpos{figure}{c}

\usepackage{titlesec}
\titlespacing\section{0pt}{12pt plus 3pt minus 3pt}{1pt plus 1pt minus 1pt}
\titlespacing\subsection{0pt}{10pt plus 3pt minus 3pt}{1pt plus 1pt minus 1pt}
\titlespacing\subsubsection{0pt}{8pt plus 3pt minus 3pt}{1pt plus 1pt minus 1pt}

\usepackage{graphics}
\usepackage{hyperref}
\usepackage{color}
\usepackage{multirow}
\usepackage{hhline}
\usepackage{subfigure}
\usepackage{lipsum}
\usepackage{flushend}
\usepackage{dblfloatfix}

\title{Frequency Enhancement for Image Demosaicking}

\usepackage{authblk}

\author{Jingyun Liu}
\author{Daiqin Yang*}
\author{Zhenzhong Chen}
\affil{School of Remote Sensing and Information Engineering,  Wuhan University}

\begin{document}

\twocolumn[ 
  \begin{@twocolumnfalse} 
  
\maketitle

\begin{abstract}

Recovering high-frequency textures in image demosaicking remains a challenging issue. 
While existing methods introduced elaborate spatial learning methods, they still exhibit limited performance.
To address this issue, a frequency enhancement approach is proposed. Based on the frequency analysis of color filter array (CFA)/demosaicked/ground truth images, we propose Dual-path Frequency Enhancement Network (DFENet), which reconstructs RGB images in a divide-and-conquer manner through fourier-domain frequency selection. In DFENet, two frequency selectors are employed, each selecting a set of frequency components for processing along separate paths. One path focuses on generating missing information through detail refinement in spatial domain, while the other aims at suppressing undesirable frequencies with the guidance of CFA images in frequency domain. Multi-level frequency supervision with a stagewise training strategy is employed to further improve the reconstruction performance. With these designs, the proposed DFENet outperforms other state-of-the-art algorithms on different datasets and demonstrates significant advantages on hard cases. 
Moreover, to better assess algorithms' ability to reconstruct high-frequency textures, a new dataset, LineSet37, is contributed, which consists of 37 artificially designed and generated images. These images feature complex line patterns and are prone to severe visual artifacts like color moir\'e after demosaicking. Experiments on LineSet37 offer a more targeted evaluation of performance on challenging cases. The code and dataset are available at https://github.com/VelvetReverie/DFENet-demosaicking.

\end{abstract}

\vspace{0.4cm}

  \end{@twocolumnfalse} 
] 

\newcommand\blfootnote[1]{%
\begingroup
\renewcommand\thefootnote{}\footnote{#1}%
\addtocounter{footnote}{-1}%
\endgroup
}

{\blfootnote{Corresponding author: Daiqin Yang, E-mail: dqyang@whu.edu.cn}}

\section{INTRODUCTION}

\begin{figure}[!b]
	\centering
	\includegraphics[width=\linewidth]{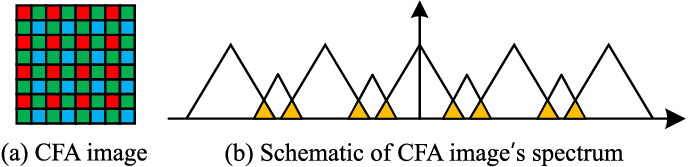}
	\caption{
		(a) CFA image under the Bayer pattern.
		(b) Schematic diagram of CFA image's spectrum. Replicas of the luminance spectrum (large triangle) do not overlap with each other but with chrominance components (small triangle). 
	}\label{fig:bayer}
\end{figure}

Most digital cameras capture images through a CCD/CMOS sensor, above which a color filter array (CFA) is placed to record a single color for each pixel. The most common configuration of the sensors follows the Bayer CFA \cite{bayer}, where there are two green components in quincunx, one red component, and one blue component in a group of four pixels, as shown in Fig. \ref{fig:bayer} (a). To obtain a full-resolution RGB image, demosaicking will be performed to interpolate the missing components from a CFA image.

Traditional methods \cite{colordifference,hirakawa,successive,lmmse,multidirectional,multi2} mostly exploited cross-color correlation and preserved the edge and texture details by leveraging spatial information from different directions. Several work \cite{buades, duran} explored image self-similarity to tackle a scenario where colors are less correlated. Moreover, based on the observation that the frequency spectrum of the CFA image is equivalent  to the combination of chrominance and luminance components, as shown in Fig. \ref{fig:bayer} (b), a few work \cite{fft1,fft2,fft3} proposed to separate them directly in frequency domain, and attributed various artifacts to improper separation. Additionally, various image priors were explored in \cite{admm, flexisp} to jointly consider demosaicking and denoising.
In recent years, deep learning-based methods have achieved excellent performance in image demosaicking \cite{deepjoint,sgnet,wildjdd,mgcc,sanet}.
They normally involve elaborate spatial designs but still exhibit limited capabilities for reconstructing high-frequency textures, resulting in false patterns such as color moir\'e.

\begin{figure}[!t]
	\centering
	\includegraphics[width=\linewidth]{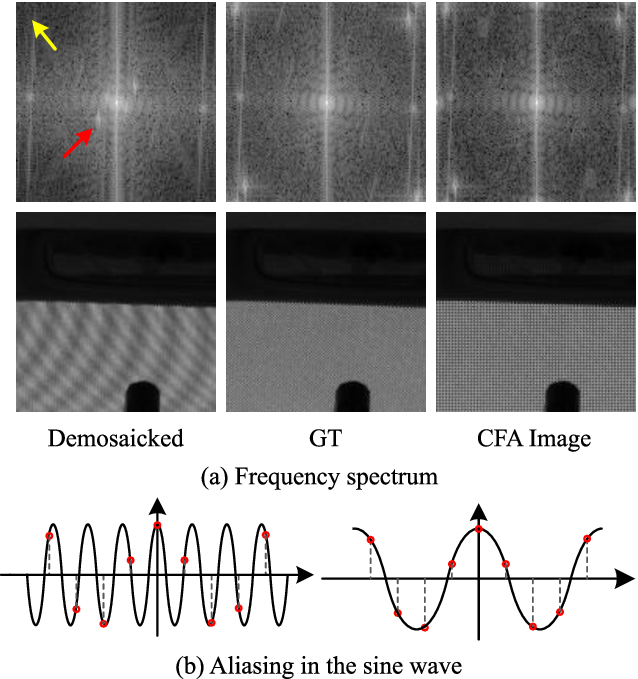}
	
	\caption{
		(a) Visualization of the demosaicked image produced from FlexISP, the ground truth image, the CFA image and their frequency spectrum. In comparison with the CFA image and the ground truth image, there are missing high frequencies (indicated by yellow arrow) and undesirable low frequencies (indicated by red arrow) in the spectrum of demosaicked image. (b) Left is the original signal, and right is the aliased signal. Red circles represent the sampling point. With a sampling rate lower than Nyquist frequency, the high-frequency signal is incorrectly reconstructed into a lower frequency, which explains the errors in the demosaicked image.}
	\label{fig:intro}
\end{figure}

Inadequate sampling rate of individual colors in the CFA image will generate frequency errors in the demosaicked RGB image due to Nyquist-Shannon sampling theorem \cite{nyquist}, including both fake low frequency and missing high frequency.
Comparing the green channel of the demosaicked image by FlexISP \cite{flexisp} and its ground truth image as shown in Fig. \ref{fig:intro} (similar errors can be observed across three color channels, thus the analysis is given based on the visualization of a single color channel), two types of errors can be found in the spectrum of the demosaicked image: undesirable low frequencies and missing high frequencies. Both the errors are manifestation of aliasing, which occurs when the sampling rate is lower than Nyquist frequency, and high-frequency information will be incorrectly aliased into lower frequency as shown in Fig. \ref{fig:intro}(b). 
\textbf{For the undesirable low frequencies}, in most cases, such frequencies are not present in the CFA image, as shown in Fig. \ref{fig:intro}(a). 
With a higher overall sampling rate of the CFA image, high frequency components of luminance can be preserved in the spectrum of CFA image without generating aliasing to the low frequency part when Nyquist frequency is achieved, as shown in Fig. \ref{fig:bayer} (b).
During the demosaicking process, such erroneous frequencies will occur if three colors are not properly separated. Therefore, under the guidance of CFA's spectrum, undesirable frequencies may be detected and eliminated. 
\textbf{For the missing high frequencies}, 
though such frequencies are preserved at the corners in the spectrum of the CFA image, it is difficult to reconstruct them directly in frequency domain, as they are severely interfered by the chrominance component \cite{fft1}, as shown in Fig. \ref{fig:bayer} (b).
According to the above analysis, different processing approaches in different domains should be employed for the two types of errors to simultaneously suppress unwanted frequencies and generate missing frequencies.

With the above findings, an image demosaicking algorithm, Dual-path Frequency Enhancement Network (DFENet), is proposed in this paper, which gives emphasized processes to two parts of selective frequencies in different manners to both alleviate distortions and recover high-frequency details.
In DFENet, two frequency selectors are introduced to respectively pass frequencies that require missing information recovering and those contain undesirable frequencies through two distinct paths. For the former, a series of detail refinement operations are performed in spatial domain to enhance the spatial learning of high frequencies. For the latter, fourier-domain masks are generated to suppress unwanted information with the guidance of the input CFA image. Moreover, multi-level frequency supervision is employed with a stage-wise training strategy to further ensure correct spectrum restoration of the demosaicked images. In the pipeline of DFENet, superior results can be achieved with simple building blocks.

In commonly used datasets, challenging structures that result in severe visual artifacts constitute only a small fraction. Therefore, experiments on these datasets struggle to distinguish between algorithms, and score differences on these datasets are often attributed to subtle variations imperceptible to the human eye.
To address this issue, a new dataset, LineSet37, is proposed to better evaluate the performance of different demosaicking algorithms on hard cases. The proposed dataset consists of 37 artificially designed and generated images with challenging line patterns that are highly susceptible to severe artifacts. 
Evaluations on LineSet37 can provide a more intuitive assessment of algorithms' effectiveness in reconstructing difficult structures compared to benchmark datasets. 


\begin{figure*}[!htbp]
	\centering
	\includegraphics[width=\linewidth]{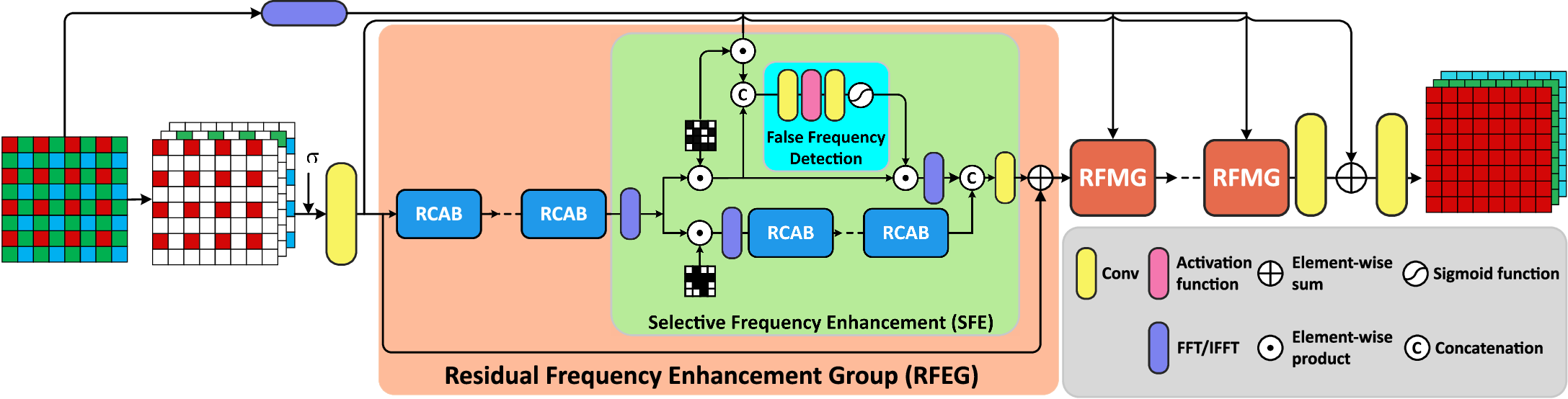}
	\caption{The architecture of the proposed Dual-path Frequency Enhancement Network (DFENet). The network comprises $M$ cascaded Residual Frequency Enhancement Groups (RFEG). In RFEG, the features first go through $N_1$ Residual Channel Attention Blocks (RCAB) for coarse feature extraction and then into the Selective Frequency Enhancement (SFE) for refinement. A divide-and-conquer manner is employed through fourier-domain frequency selection to simultaneously generate missing details (the lower path in SFE) and suppress false frequencies (the upper path in SFE) under the guidance of the input CFA image. The refined features from the two paths are fused with a single convolution layer.}
	\label{fig:framework}
\end{figure*}

Overall, the contributions of the paper are as follows:

\begin{itemize}
    \item A Dual-path Frequency Enhancement Network (DFENet) is proposed for image demosaicking, which provides enhanced ability of reconstructing high-frequency textures and alleviating color distortion. With fourier-domain frequency selection, it can simultaneously generate missing high frequencies and suppress fake low frequencies to ensure precise texture reconstruction during demosaicking.

    \item A new dataset, LineSet37, is contributed with artificially designed and generated images to evaluate reconstruction of challenging structures. Images in LineSet37 feature complex line patterns where severe artifacts are prone to occur after image demosaicking. Evaluation on this dataset provides a more targeted reflection of the algorithms' effectiveness on hard cases.
    
    \item We demonstrate that by processing selective frequency components in a divide-and-conquer manner, the proposed method can achieve superior performance on benchmark datasets while show significant advantages on challenging cases, both in quantitative metrics and perceptual quality.

\end{itemize}

The remainder of the paper is organized as follows. First, Section \ref{related} reviews research work on image demosaicking. Then, Section \ref{methods} presents the detail of the proposed DFENet. After that, Section \ref{dataset} describes the proposed dataset LineSet37. Experiment settings and results are given in Section \ref{experiments}. Finally, Section \ref{conclusion} makes a conclusion for our work.

\section{Related Work}\label{related}

Traditional methods mostly inferred the missing colors from the neighboring pixels or recorded colors. 
As the difference of colors are normally low-pass signals, Pei \textit{et al.} \cite{colordifference} proposed to interpolate the color difference instead of directly estimating R, G and B values, improving image quality by reducing hue artifacts. Based on the color difference rule, many work explored different interpolation and estimation approaches.
Hirakawa \textit{et al.} \cite{hirakawa} alleviated color artifacts by performing directional interpolation under the guidance of homogeneity map.
Li \cite{successive} proposed an iterative demosaicking algorithm, attributing the aliasing effect to not sufficiently enforcing the color difference rule.
Zhang \textit{et al.} \cite{lmmse} avoided the interpolation errors introduced by gradient-guided approach and achieved robust demosaicking results by using linear minimum mean square-error estimation (LMMSE) to estimate color difference signals in horizontal and vertical directions, and further fusing them with a weighted average strategy. 
Additionally, Chen \textit{et al.} \cite{multidirectional} achieved greater edge detection power by interpolating in eight directions, and further reduced color artifacts using a realigning postprocessing method with an antialiasing fixed-tap finite impulse response filter.
Wang \textit{et al.} \cite{multi2} used multi-directional interpolation to recover green channel and estimated the other two channels using guided filter, which effectively preserved edges. 
Besides, several works tackled a scenario where there are more saturated regions and colors are less correlated. 
Buades \textit{et al.} \cite{buades} utilized image self-similarity to alleviate artifacts like zipper artifacts.
To adapt to different degrees of color correlation, Duran \textit{et al.} \cite{duran} performed directional interpolation based on chromatic smoothness and applied non-local means algorithm \cite{nonlocal} on color differences. 
Moreover, a few works were based on fourier analysis \cite{fft1, fft2, fft3}. 
Alleysson \textit{et al.} \cite{fft1} discovered the luminance and chrominance components overlap in the fourier domain, attributing the appearance of aliasing to unsuccessfully decoupling luminance and chrominance, and de-multiplexed luminance and chrominance in spatial domain.
Based on the discovery, the authors \cite{fft2} further proposed to estimate suitable filters to separate the two components in the fourier domain.
Observing that vertical and horizontal frequencies of a specific chrominance component differently overlap with luminance, Dubois \cite{fft3} estimated filters for both directions and adaptively fused the filtered results, effectively reducing spectral crosstalk. 
Moreover, Heide \cite{flexisp} proposed an end-to-end pipeline to jointly consider demosaicking, denoising and other tasks using natural image priors.
Tan \textit{et al.} \cite{admm} improved demosaicking and denoising performance and robust by incorporating various effective priors like smoothness prior and edge-preserving prior.

Recently, deep learning-based methods have achieved excellent performance in various low-level vision tasks, including demosaicking. Gharbi \textit{et al.} \cite{deepjoint} first introduced convolutional neural network into this task, and collected a dataset of hard patches with luminance artifacts and color moir\'e. In addition, many authors exploited inter-color correlation. Liu \textit{et al.} \cite{sgnet} initially reconstructed the green color channel and guided the rest of the color channels under the guidance of the green channel. 
Chen \textit{et al.} \cite{wildjdd} identified the limitation caused by imperfect reference images, and improved network's performance under ground truth certainty by formulating a two-stage data degradation and optimizing the network with an evidence lower bound.
Zhang \textit{et al.} \cite{mgcc} utilized the strengths of all color channels and developed a three-branch network to achieve mutual guidance between three color channels. Moreover, mosaic pattern was introduced to fulfill spatial adaptive information integration \cite{sanet}. 
Guan \textit{et al.} \cite{deformable} introduced deformable convolution into demosaicking, and proposed several strategies like offset sharing to reduce the computational cost and memory requirement of deformable convolution.
Elgendy \textit{et al.} \cite{quad2} performed demosaicking for low-light color images with learnable filters to separate luminance and chrominance in frequency domain. 
Xing \textit{et al.} \cite{xing} proposed a network for joint image demosaicking, denoising and super resolution.
Bai \textit{et al.} \cite{freq2} utilized frequency information to obtain global information, and performed frequency selection to recover luminance and chrominance components, which leveraged the periodicity of CFA images.
Moreover, to balance computational cost and image quality, Ma \textit{et al.} \cite{search} firstly proposed to synthesize demosaicking pipeline with a multi-objective, discrete-continuous search. Lee \textit{et al.} \cite{quad4} presented an efficient unified model that can process both Bayer and non-Bayer CFA images.

\section{The Proposed Method}\label{methods}

In this section, we reveal the details of the proposed Dual-path Frequency Enhancement Network (DFENet) and the training strategy.

\subsection{The Overall Architecture}

The overall pipeline of the proposed DFENet is illustrated in Fig. \ref{fig:framework}. It consists of $M$ repeated Residual Frequency Enhancement Groups (RFEG), each group containing total $N$ Residual Channel Attention Blocks (RCAB) \cite{rcan}. RCAB is used because it can exploit inter- and intra-color correlation.

\begin{figure}[t]
	\centering
	\includegraphics[width=\linewidth]{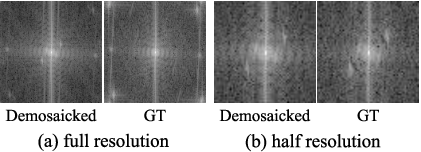}
	\caption{Comparison of frequency spectrum between full-resolution and half-resolution image. Spectrums in (a) respectively correspond to the demosaicked image and the ground truth image, and those in (b) correspond to the downscaled demosaicked image and ground truth image with pixelunshuffle. The other three components obtained from pixelunshuffle are not shown as they are similar to the one presented.}
	\label{fig:input}
\end{figure}

DFENet takes the noisy CFA image $ I_\mathrm{CFA}\in \mathbb{R}^{H\times W} $ as input, with each pixel recording a single color. $H$, $W$ represent the height and width of the input. Due to the spatial-agnostic behavior of convolution \cite{involution}, directly applying convolution to $I_\mathrm{CFA}$ is unsuitable as sliding windows at different spatial locations involve distinct arrangements of three colors. Hence, to better extract information from each color channel, $I_\mathrm{CFA}$ is first rearranged into three-channel format $I_\mathrm{input}\in\mathbb{R}^{3\times H\times W}$, where values of unrecorded color components are set to zero and color-specific kernels can focus on capturing single-channel features. To make the network aware of the input's noise level, $I_\mathrm{input}$ is concatenated with a noise map $I_\mathrm{noise}\in \mathbb{R}^{1\times H\times W}$ with all pixel values assigned to $\sigma$ before feeding into the network, where $\sigma$ denotes the noise level of the input. 
It is noted that in most demosaicking networks, CFA images will be initially packed into a half-resolution four-channel RGGB format and upscaled to the original resolution with pixelshuffle in the end. However, when fourier-domain processing is required, this approach is unsuitable, as employing the RGGB format processes the image at half resolution, which will diminish the discrepancies between the spectrum of the demosaicked image and the ground truth image, as shown in Fig. \ref{fig:input}. It is because the patterns are blurred in the downscaled image, where aliasing effects are reduced. Therefore, full-resolution CFA image is used here.

\begin{table}
	\renewcommand{\arraystretch}{1.1}
	\centering
	\caption{Notation Conventions}
	\begin{tabular}{c|p{6.5cm}}
		\hline
		\textbf{Symbol} & \textbf{Definitions} \\
		\hline
		$F$ & Feature map in spatial domain \\
		\hline
		$\mathcal{F}$ & Fourier representation of $F$ \\
		\hline
		\multirow{2}{*}{$F_{\mathrm{gn}}$} & Output feature from the path that focuses on generating missing details \\
		\hline
		\multirow{2}{*}{$F_{\mathrm{sp}}$} & Output feature from the path that focuses on suppressing undesirable frequencies \\
		\hline
		\multirow{2}{*}{$\mathcal{F}^{\mathrm{select}}_\mathrm{gn}$} & Selected frequency components for the path that focuses on generating missing details \\
		\hline
		\multirow{2}{*}{$\mathcal{F}^{\mathrm{select}}_\mathrm{sp}$} & Selected frequency components for the path that focuses on suppressing undesirable frequencies \\

		\hline
	\end{tabular}
\end{table}

Given the input $[I_\mathrm{input}, I_\mathrm{noise}]$, a $3\times 3$ convolution $\mathrm{W}_\mathrm{init}$ is first applied to obtain initial features $F_0\in \mathbb{R}^{C\times H\times W}$: 
\begin{equation}
    F_0 = \mathrm{W_{init}}([I_\mathrm{input}, I_\mathrm{noise}]),
\end{equation}
where $C$ is the number of channels.
$F_0$ will then progressively go through $M$ RFEGs and the output feature of the $i$-th RFEG is denoted as $F^i\in \mathbb{R}^{C\times H\times W}$. The input of the $i$-th RFEG, which is the output features of the previous residual group $F^{i-1}$, will first go through $N_1$ RCABs, obtaining coarse features $F^i_\mathrm{coarse}$, which can be calculated as:
\begin{equation}
	F_\mathrm{coarse}^i = \mathrm{RB}_{N_1}(\ldots \mathrm{RB}_{1}(F^{i-1})).
\end{equation}
Then, in Selective Frequency Enhancement (SFE) module, 
two frequency selectors $S^i_\mathrm{gn}\in \mathbb{R}^{1\times H\times W}$ and $S^i_\mathrm{sp}\in \mathbb{R}^{1\times H\times W}$ are introduced for the two paths that respectively focus on generating missing information and suppressing undesirable frequencies, obtaining selected frequency components $\mathcal{F}^{i,\mathrm{select}}_\mathrm{gn}\in \mathbb{R}^{C\times H\times W}$ and $\mathcal{F}^{i,\mathrm{select}}_\mathrm{sp}\in \mathbb{R}^{C\times H\times W}$ from the Fast Fourier Transformation (FFT) results of $F_\mathrm{coarse}^i$. The process can be formulated as:
\begin{equation}
	\mathcal{F}_\mathrm{gn}^{i,\mathrm{select}} = S^i_\mathrm{gn}\odot \mathrm{FFT}(F_\mathrm{coarse}^i)
\end{equation}
\begin{equation}
 \mathcal{F}_\mathrm{sp}^{i,\mathrm{select}} = S^i_\mathrm{sp}\odot \mathrm{FFT}(F_\mathrm{coarse}^i)
\end{equation}
where $\odot$ denotes the Hadamard product.
$\mathcal{F}_\mathrm{gn}^{i,\mathrm{select}}$, which are frequencies that suffer from information loss, will be transformed back into spatial domain to go through another $N_2$ RCABs for further spatial detail refinement. $\mathcal{F}_\mathrm{sp}^{i,\mathrm{select}}$, which are frequencies where undesirable frequencies are located, will go through the frequency suppressor to reduce those false frequencies with the guidance of CFA image' frequency spectrum. 
The outputs of the two paths $F^i_\mathrm{gn}$ and $F^i_\mathrm{sp}$ are then concatenated and fused with a convolution, obtaining the output of the $i$-th residual group $F^i$, which can be calculated as:
\begin{equation}\label{eqa:fuse}
    F^i = \mathrm{W}_\mathrm{fuse}([F^i_\mathrm{gn}, F^i_\mathrm{sp}]),
\end{equation}
where $\mathrm{W}_\mathrm{fuse}$ is a $3\times 3$ convolution used to fuse the features from the two paths. 

The final demosaicked output will be generated from progressively refined features with a $3\times 3$ convolution $\mathrm{W_{final}}$:
\begin{equation}
    I_\mathrm{dm} = \mathrm{W_{final}}(F^M).
\end{equation}

\subsection{Selective Frequency Enhancement}

The proposed Selective Frequency Enhancement (SFE) enhances the ability of reconstructing challenging structures via a divide-and-conquer approach, where selected frequencies will undergo different processing, including both generating missing high frequencies and suppressing undesirable frequencies. 

To select those frequencies, two sets of learnable frequency selectors $S^i_\mathrm{gn}$ and $S^i_\mathrm{sp}$ are introduced, where the positions of the selected frequencies are assigned to 1 while the others are assigned to 0. Instead of directly learning the selectors at size $(H, W)$, low-resolution selectors $S^\mathrm{lr}_\mathrm{gn}\in \mathbb{R}^{1\times \frac{H}{s}\times \frac{W}{s}}$ and $S^\mathrm{lr}_\mathrm{sp}\in \mathbb{R}^{1\times \frac{H}{s}\times \frac{W}{s}}$ are used to avoid overly fragmented frequency selections, where $s$ is the downscaling factor. The two low-resolution selectors are jointly optimized with the other parameters of DFENet. When selecting frequencies, $S^\mathrm{lr}_\mathrm{gn}$ and $S^\mathrm{lr}_\mathrm{sp}$ are first upscaled with bilinear interpolation to the size of $(H,W)$ and then binarized to 1/0 with the threshold of 0. Reparameterization trick is employed to make the operation differentiable. 

The IFFT result $F^{i,\mathrm{select}}_\mathrm{gn}$ of the selected frequency component $\mathcal{F}^{i,\mathrm{select}}_\mathrm{gn}$ will go through $N_2$ RCABs, enabling the network to concentrate on refining complex structural information. The refinement is performed in spatial domain rather than the fourier domain because adjustment of texture details is more likely to be localized while each frequency component is a reflection of global information. The process of this path can be formulated as:
\begin{equation}
	F^i_\mathrm{gn} = \mathrm{RB}_{N_1+N_2}(\ldots \mathrm{RB}_{N_1+1}(F^{i,\mathrm{select}}_\mathrm{gn})),
\end{equation}
In contrast, undesirable frequencies are removed in the fourier domain with the guidance of the CFA image $I_\mathrm{CFA}$. The fourier representation of $I_\mathrm{CFA}$ is first multiplied with the frequency selector $S^i_\mathrm{sp}$to obtain $\mathcal{F}^\mathrm{select}_\mathrm{CFA}$:
\begin{equation}
	\mathcal{F}^{i,\mathrm{select}}_\mathrm{CFA} = S^i_\mathrm{sp}\odot \mathrm{FFT}(I_\mathrm{CFA})
\end{equation}
The selective frequencies $\mathcal{F}^{i,\mathrm{select}}_\mathrm{CFA}$ and $\mathcal{F}^{i,\mathrm{select}}_\mathrm{sp}$ will be then fed into the False Frequency Detection (FFD) module to generate frequency suppressors. By learning the differences between them, undesirable frequencies can be located. Specifically, the real and imaginary parts of $\mathcal{F}^{i,\mathrm{select}}_\mathrm{CFA}$ and $\mathcal{F}^{i,\mathrm{select}}_\mathrm{sp}$ are first concatenated, and differences are learned through two $1\times 1$ convolutions with an activation function. The frequency suppressor is generated with a sigmoid function, mapping the values of differences to the range of [0,1],indicating the extent ot which the frequency should be preserved. Undesirable frequencies can be reduced by multiplying the frequency suppressor with $\mathcal{F}^{i,\mathrm{select}}_\mathrm{sp}$.
The formulation of this path is presented:
\begin{equation}
    F^i_\mathrm{sp} = \mathrm{IFFT}(\mathcal{F}^{i,\mathrm{select}}_\mathrm{sp}\odot \mathrm{M_{sp}}([\mathcal{F}_\mathrm{sp}^{i,\mathrm{select}}, \mathcal{F}^{i,\mathrm{select}}_\mathrm{CFA}])),
\end{equation}
where $\mathrm{M_{sp}}$ represents the process to generate frequency suppressor.
At the end of each RFEG, the refined features from the two paths are fused with a $3\times 3$ convolution as formulated in Equation \ref{eqa:fuse}.

\subsection{Optimization}

The network is optimized with two losses: reconstruction loss and multi-level frequency loss. The reconstruction loss is calculated by L1 distance between the final output $I_\mathrm{dm}$ and the ground truth $I_\mathrm{gt}$:
\begin{equation}
    L_\mathrm{rec} = \|I_\mathrm{dm} - I_\mathrm{gt}\|
\end{equation}
For the multi-level frequency loss, the intermediate feature outputs of $M$ residual groups are supervised by $I_\mathrm{gt}$. To compute the loss, extra restorers $\mathrm{W}^i_\mathrm{st}$ implemented by $1\times 1$ convolution are introduced to map the features $F^i$ to the RGB domain, obtaining $I^i = \mathrm{W}^i_\mathrm{st}(F^i)$. Then the distance is computed between the FFT results of $I^i$ and $I_\mathrm{gt}$:
\begin{equation}
    L_\mathrm{fft} = \sum^M_{i=1}\|\mathrm{FFT}(I^i) - \mathrm{FFT}(I_\mathrm{gt})\|, 
\end{equation}
The overall training loss can be formulated as:
\begin{equation}
    L = L_\mathrm{rec} + \lambda L_\mathrm{fft},
\end{equation}
where $\lambda$ is the trade-off that balances the two losses and is set to 0.01.

To better supervise the frequency learning of the network, a stagewise training strategy is employed to emphasize the multi-level frequency supervision, where backward propagation is performed twice within a mini-batch respectively using $L_\mathrm{fft}$ and $L_\mathrm{rec}$. Specifically, in the first stage, the parameters of restorers and DFENet are optimized by minimizing $\lambda L_\mathrm{fft}$. In the second stage, the parameters of DFENet are optimized by minimizing $L_\mathrm{rec}$.

\section{The Proposed Dataset LineSet37}\label{dataset}

\begin{figure}
	\centering
	\includegraphics[width=\linewidth]{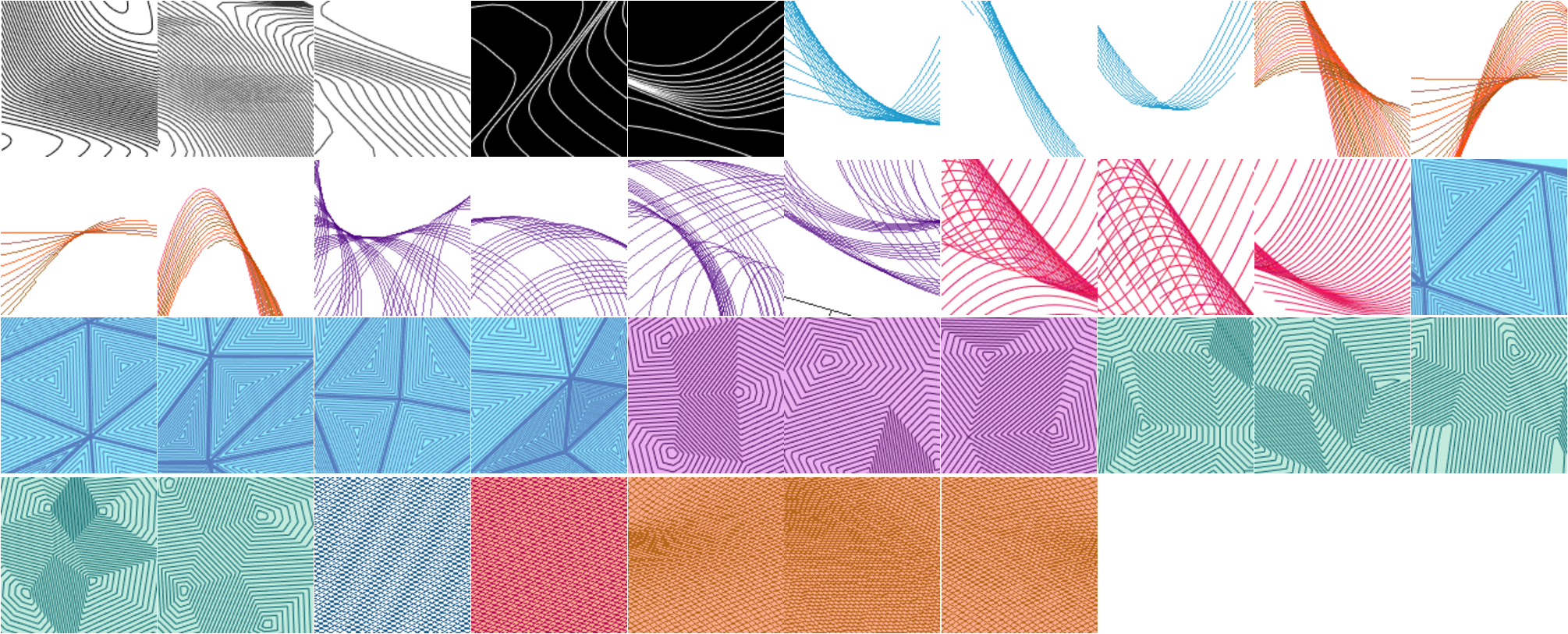}
	\caption{Images from the proposed dataset LineSet37.}
	\label{fig:dataset}
\end{figure}

Benchmark datasets for image demosaicking, such as Kodak, Set14, Urban100 and MIT moir\'e, cover a wide range of scenarios. However, only a small fraction of these represent hard cases prone to perceptible visual artifacts, while the majority are less susceptible. As a result, experiments on these datasets struggle to differentiate the performance of various algorithms, and score differences often arise from subtle discrepancies that are imperceptible to the human eyes.
Though MIT moir\'e consists of hard cases with luminance artifacts and color moir\'e, images that will suffer from severe distortion after demosaicking only constitute a small proportion. To better evaluate the capability of different algorithms, the LineSet37 dataset is proposed with complex line patterns, which are illustrated in Fig. \ref{fig:dataset}. The patterns in the proposed dataset are artificially designed and generated. Complex dense curves such as contour lines, trigonometric functions and repetitive nested polygons are plotted in three-dimensional space. By applying different colors to the lines and the background and rotating the viewing angle, several patterns prone to artifacts are generated. LineSet37 consists of 37 images with resolution of $128\times 128$. CFA images can be synthesized by sampling these images according to the Bayer pattern. Gaussian noises are added for noisy input with noise levels of 5, 10 and 15, following \cite{sgnet}. After the demosaicking process, the reconstruction results of images in LineSet37 suffer from large areas of distortions. Various severe color distortions can be observed in the figure. Therefore, the LineSet37 dataset can provide a more targeted assessment of algorithms' performance on challenging structures.

\begin{table*}[!t]
	\centering
	\caption{Quatitative results (PSNR/SSIM/IW-SSIM/LPIPS) on four benchmark datasets: Kodak, Set14, Urban100, and MIT moir\'e. Results with the best performance are highlighted in \textbf{bold}. '-' indicates that the results are not provided by the corresponding paper.}
	\label{tab:comparison}
	\scriptsize
	
	\begin{tabular}{cccccc}
		\toprule
		\multirow{2}{*}{Methods} & \multirow{2}{*}{$\sigma$}  & Kodak & Set14 & Urban100 & MIT moir\'e \\
		\cmidrule{3-6}
		&& PSNR/SSIM/IW-SSIM/LPIPS & PSNR/SSIM/IW-SSIM/LPIPS & PSNR/SSIM/IW-SSIM/LPIPS & PSNR/SSIM/IW-SSIM/LPIPS\\
		\midrule
		ADMM\cite{admm} & \multirow{12}{*}{0} &  31.87/0.8871/0.8495/0.1032 & 30.98/0.8854/0.8498/0.0990 & 28.75/0.8926/00.8534/0.0711 & 28.52/0.8338/0.7919/0.1009 \\
		FlexISP\cite{flexisp} && 35.32/0.9422/0.9246/0.0522 & 33.51/0.9233/0.8986/0.0672 & 34.22/0.9504/0.9392/0.0309 & 32.07/0.9051/0.8806/0.0587 \\
		Deepjoint\cite{deepjoint} &&  40.45/0.9833/0.9820/0.0102 & 38.92/0.9758/0.9701/0.0088 & 36.53/0.9749/0.9700/0.0095 & 35.08/0.9473/0.9347/0.0181\\
		Kokkinos\cite{deepunfold} &&  38.81/0.9722/0.9703/0.0195 & 36.76/0.9473/0.9370/0.0249 & 35.89/0.9639/0.9594/0.0180 & 32.52/0.8947/0.8753/0.0543\\
  		SGNet\cite{sgnet} &&  - & - & - & - \\
  		Wild-JDD\cite{wildjdd} && - & - & - & - \\
  		Wild-JDD*\cite{wildjdd} &&  - & - & - & - \\
		ConvIR\cite{convir} && 42.80/0.9897/0.9890/0.0035 & 42.73/0.9869/0.9846/0.0031 & 38.95/0.9822/0.9765/0.0043 & 35.50/0.9361/0.9222/0.0266 \\
		FSNet\cite{fsnet} &&  42.74/0.9896/0.9889/0.0033 & 42.85/0.9876/0.9854/0.0030 & 38.88/0.9821/0.9764/0.0041 & 35.56/0.9362/0.9229/0.0266 \\
		GRL\cite{grl} &&  42.35/{0.9905}/{0.9899}/{0.0030} & 43.70/0.9883/0.9864/0.0027 & 40.23/{0.9859}/{0.9816}/{0.0030} & {37.25}/{0.9440}/{0.9328}/0.0242 \\
		FourierISP\cite{fourierisp} &&  40.61/0.9817/0.9811/0.0079 & 39.72/0.9780/0.9751/0.0092 & 36.97/0.9735/0.9692/0.0092 & 35.08/0.9375/0.9238/\textbf{0.0204}\\
		SANet\cite{sanet} &&  42.40/0.9893/0.9883/0.0034 & 42.43/0.9871/0.9846/0.0033 & 38.29/0.9809/0.9743/0.0046 & 34.33/0.9303/0.9137/0.0275 \\
		MGCC\cite{mgcc} && 42.38/0.9893/0.9885/0.0035 & 42.20/0.9870/0.9852/0.0031 & 38.83/0.9826/0.9778/0.0039 & 35.99/0.9397/0.9275/0.0255 \\
		DFENet (Ours) &&  \textbf{43.39}/\textbf{0.9905}/\textbf{0.9899}/\textbf{0.0027} & \textbf{44.74}/\textbf{0.9893}/\textbf{0.9875}/\textbf{0.0024} & \textbf{41.25}/\textbf{0.9871}/\textbf{0.9832}/\textbf{0.0025} & \textbf{37.82}/\textbf{0.9461}/\textbf{0.9362}/0.0230 \\
		\midrule
		
		ADMM\cite{admm} & \multirow{12}{*}{5} &  31.90/0.8785/0.8482/0.1152 & 31.12/0.8772/0.8504/0.1045 & 29.04/0.8863/0.8596/0.0731 & 28.63/0.8250/0.7913/0.1099 \\
		FlexISP\cite{flexisp} & &  31.45/0.8686/0.8459/0.0752 & 30.84/0.8681/0.8386/0.0788 & 30.47/0.8908/0.8803/0.0490 & 29.25/0.8518/0.8244/0.0742 \\
		Deepjoint\cite{deepjoint} &&  36.23/0.9444/0.9413/0.0327 & 35.06/0.9346/0.9245/0.0423 & 34.03/0.9465/0.9460/0.0189 & 32.81/0.9168/0.9045/0.0296 \\
		Kokkinos\cite{deepunfold} &&  35.26/0.9324/0.9265/0.0324 & 33.93/0.9198/0.9058/0.0256 & 33.51/0.9398/0.9351/0.0233 & 30.74/0.8690/0.8507/0.0586 \\
		SGNet\cite{sgnet} &&  - & - & 34.54/0.9533/\quad-\quad/0.0299 & 32.15/0.9043/\quad-\quad/0.0691 \\
		Wild-JDD\cite{wildjdd} && 36.88/0.9520/\quad-\quad/ \quad-\quad\quad & - & 34.70/0.9534/\quad-\quad/ \quad-\quad\quad & 32.29/0.8987/\quad-\quad/ \quad-\quad\quad \\
		Wild-JDD*\cite{wildjdd} &&  36.97/0.9526/\quad-\quad/ \quad-\quad\quad & - & 34.83/0.9540/\quad-\quad/ \quad-\quad\quad  & 32.39/0.8999/\quad-\quad/ \quad-\quad\quad \\
		ConvIR\cite{convir} &&  37.12/0.9522/0.9495/0.0252 & 36.21/0.9424/0.9346/0.0400 & 35.24/0.9528/0.9533/0.0144 & 32.95/0.9080/0.8955/0.0349 \\
		FSNet\cite{fsnet} && 37.10/0.9525/0.9499/0.0233 & 36.36/0.9441/0.9370/0.0368 & 35.17/0.9532/0.9534/0.0126 & 32.95/0.9079/0.8958/0.0345 \\
		GRL\cite{grl} &&  {37.38}/{0.9537}/{0.9512}/0.0239 & {36.47}/0.9442/0.9373/0.0376 & {35.97}/{0.9575}/{0.9590}/0.0123 & {34.08}/{0.9188}/{0.9093}/0.0311 \\
		FourierISP\cite{fourierisp} && 36.22/0.9420/0.9409/\textbf{0.0183} & 35.58/0.9384/0.9314/\textbf{0.0315} & 34.32/0.9443/0.9454/0.0143 & 32.74/0.9100/0.8977/\textbf{0.0249} \\
		SANet\cite{sanet} &&  36.98/0.9516/0.9485/0.0242 & 36.14/0.9426/0.9347/0.0386 & 34.88/0.9519/0.9510/0.0139 & 32.32/0.9032/0.8884/0.0361 \\
		MGCC\cite{mgcc} &&  37.15/0.9531/0.9503/0.0244 & 36.19/0.9444/0.9367/0.0381 & 35.44/\textbf{0.9597}/0.9559/0.0128 & 33.37/0.9168/0.9029/0.0329 \\
		DFENet (Ours) && \textbf{37.45}/\textbf{0.9543}/\textbf{0.9518}/0.0224 & \textbf{36.65}/\textbf{0.9461}/\textbf{0.9399}/0.0355 & \textbf{36.24}/0.9585/\textbf{0.9604}/\textbf{0.0115} & \textbf{34.37}/\textbf{0.9221}/\textbf{0.9139}/0.0299  \\
		\midrule
		
		ADMM\cite{admm} & \multirow{12}{*}{10} &  31.29/0.8593/0.8289/0.1314 & 30.53/0.8475/0.8341/0.1130 & 28.98/0.8725/0.8563/0.0766 & 28.32/0.8043/0.7775/0.1232 \\
		FlexISP\cite{flexisp} & &  28.84/0.7580/0.7349/0.1270 & 28.27/0.7779/0.7413/0.1014 & 27.62/0.7967/0.7906/0.0864 & 26.84/0.7682/0.7383/0.0953 \\
		Deepjoint\cite{deepjoint} &&  33.26/0.9005/0.8900/0.0646 & 32.58/0.8895/0.8745/0.0880 & 31.71/0.9110/0.9129/0.0348 & 30.56/0.8718/0.8593/0.0526 \\
		Kokkinos\cite{deepunfold} &&  30.94/0.8278/0.8118/0.1069 & 31.33/0.8647/0.8418/0.0501 & 30.23/0.8728/0.8688/0.0588 & 28.12/0.8002/0.7797/0.0902 \\
		SGNet\cite{sgnet} &&  - & - & 32.14/0.9229/\quad-\quad/0.0546 & 30.09/0.8619/\quad-\quad/0.1034 \\
		Wild-JDD\cite{wildjdd} &&  33.81/0.9127/\quad-\quad/ \quad-\quad\quad & - & 32.42/0.9288/\quad-\quad/ \quad-\quad\quad & 30.30/0.8645/\quad-\quad/ \quad-\quad\quad \\
		Wild-JDD*\cite{wildjdd} &&  33.88/0.9136/\quad-\quad/ \quad-\quad\quad & - & 32.54/0.9299/\quad-\quad/ \quad-\quad\quad & 30.37/0.8657/\quad-\quad/ \quad-\quad\quad\\
		ConvIR\cite{convir} &&  34.01/0.9144/0.9032/0.0549 & 33.12/0.8946/0.8787/0.0936 & 32.74/0.9216/0.9251/0.0300 & 30.73/0.8690/0.8555/0.0535 \\
		FSNet\cite{fsnet} &&  34.00/0.9144/0.9032/0.0549 & 33.19/0.8950/0.8792/0.0931 & 32.71/0.9221/0.9256/0.0289 & 30.74/0.8689/0.8559/0.0521 \\
		GRL\cite{grl} &&  {34.21}/0.9161/0.9057/0.0546 & 33.46/0.8985/0.8848/0.0877 & 33.40/0.9276/0.9331/0.0272 & {31.68}/{0.8848}/{0.8753}/0.0460 \\
		FourierISP\cite{fourierisp} &&  33.41/0.9033/0.8955/\textbf{0.0359} & 33.11/0.8975/0.8845/\textbf{0.0732} & 32.16/0.9135/0.9180/\textbf{0.0227} & 30.61/0.8734/0.8612/\textbf{0.0340} \\
		SANet\cite{sanet} &&  33.88/0.9128/0.9014/0.0527 & 33.24/0.8970/0.8828/0.0886 & 32.45/0.9198/0.9220/0.0283 & 30.32/0.8635/0.8482/0.0557 \\
		MGCC\cite{mgcc} && 34.15/0.9159/0.9049/0.0541 & 33.52/\textbf{0.9021}/0.8842/0.0893 & 33.15/\textbf{0.9371}/0.9303/0.0264 & 31.17/0.8849/0.8661/0.0479 \\
		DFENet (Ours) &&  \textbf{34.31}/\textbf{0.9179}/\textbf{0.9075}/0.0498 & \textbf{33.58}/{0.9004}/\textbf{0.8878}/0.0844 & \textbf{33.67}/0.9300/\textbf{0.9359}/0.0246 & \textbf{31.89}/\textbf{0.8890}/\textbf{0.8814}/0.0443 \\
		\midrule
		
		ADMM\cite{admm} & \multirow{12}{*}{15} &  30.36/0.8367/0.8044/0.1480 & 29.55/0.8332/0.8105/0.1256 & 28.46/0.8553/0.8432/0.0840 & 27.66/0.7813/0.7570/0.1342 \\
		FlexISP\cite{flexisp} & &  26.87/0.6464/0.6222/0.2129 & 26.29/0.6827/0.6395/0.1524 & 25.66/0.7054/0.6996/0.1424 & 25.10/0.6860/0.6521/0.1279 \\
		Deepjoint\cite{deepjoint} &&  31.34/0.8583/0.8416/0.0999 & 30.75/0.8455/0.8266/0.1271 & 29.95/0.8767/0.8787/0.0534 & 28.95/0.8277/0.8144/0.0790 \\
		Kokkinos\cite{deepunfold} &&  27.18/0.6689/0.6468/0.2458 & 28.91/0.7860/0.7530/0.0995 & 26.91/0.7598/0.7496/0.1468 & 25.55/0.6981/0.6710/0.1637 \\
		SGNet\cite{sgnet} && - & - & 30.37/0.8923/\quad-\quad/0.0793 & 28.60/0.8188/\quad-\quad/0.1412 \\
		Wild-JDD\cite{wildjdd} &&  31.92/0.8765/\quad-\quad/ \quad-\quad\quad & - & 30.79/0.9055/\quad-\quad/ \quad-\quad\quad & 28.89/0.8310/\quad-\quad/ \quad-\quad\quad \\
		Wild-JDD*\cite{wildjdd} &&31.99/0.8777/\quad-\quad/ \quad-\quad\quad & - & 30.89/0.9070/\quad-\quad/ \quad-\quad\quad & 28.95/0.8325/\quad-\quad/ \quad-\quad\quad \\
		ConvIR\cite{convir} &&  32.10/0.8798/0.8606/0.0871 & 31.07/0.8527/0.9309/0.1358 & 30.97/0.8940/0.8977/0.0472 & 29.19/0.8307/0.8152/0.0768 \\
		FSNet\cite{fsnet} &&  32.08/0.8791/0.8607/0.0907 & 30.97/0.8500/0.8257/0.1396 & 30.95/0.8942/0.8982/0.0473 & 29.22/0.8316/0.8166/0.0733 \\
		GRL\cite{grl} && {32.28}/0.8818/0.8637/0.0858 & 31.51/0.8595/0.8418/0.1257 & 31.60/0.9016/0.9083/0.0434 & {30.05}/{0.8511}/{0.8404}/0.0647 \\
		FourierISP\cite{fourierisp} && 31.64/0.8695/0.8552/\textbf{0.0552} & 31.35/{0.8623}/{0.8456}/\textbf{0.1060} & 30.57/0.8872/0.8928/\textbf{0.0332} & 29.12/0.8397/0.8367/\textbf{0.0450} \\
		SANet\cite{sanet} &&  31.97/0.8776/0.8587/0.0808 & 31.28/0.8571/0.8384/0.1283 & 30.70/0.8913/0.8937/0.0441 & 28.87/0.8249/0.8077/0.0790 \\
		MGCC\cite{mgcc} &&  32.34/0.8829/0.8638/0.0836 & \textbf{31.75}/\textbf{0.8658}/0.8402/0.1291 & 31.59/\textbf{0.9169}/0.9068/0.0404 & 29.71/0.8544/0.8307/0.0656 \\
		DFENet (Ours) &&  \textbf{32.39}/\textbf{0.8851}/\textbf{0.8671}/0.0777 & 31.62/0.8617/\textbf{0.8458}/0.1225 & \textbf{31.93}/0.9061/\textbf{0.9138}/0.0382 & \textbf{30.25}/\textbf{0.8575}/\textbf{0.8490}/0.0607 \\
		\midrule
		
	\end{tabular}
\end{table*}

%

\begin{table*}[!htbp]
	\centering
	\caption{Quatitative results (PSNR/SSIM/IW-SSIM/LPIPS) on LineSet37. Results with the best and second performance are highlighted in \textbf{bold}.}
	\label{tab:comparison_textures}
	\scriptsize
	\begin{tabular}{ccccc}
		\toprule
		\multirow{2}{*}{Methods} &  $\sigma=0$ & $\sigma=5$ & $\sigma=10$ & $\sigma=15$ \\
		\cmidrule{2-5}
		& PSNR/SSIM/IW-SSIM/LPIPS & PSNR/SSIM/IW-SSIM/LPIPS & PSNR/SSIM/IW-SSIM/LPIPS & PSNR/SSIM/IW-SSIM/LPIPS\\
		\midrule
		
		ADMM\cite{admm} &  18.89/0.7595/0.7043/0.1260 & 18.99/0.7569/0.7100/0.1212 & 19.22/0.7581/0.7213/0.1132 & 19.43/0.7622/0.7309/0.1062  \\
		FlexISP\cite{flexisp} & 25.79/0.8976/0.8820/0.0391 & 25.05/0.8900/0.8794/0.0322 & 23.22/0.8459/0.8506/0.0446 & 21.70/0.8105/0.8211/0.0658 \\
		Deepjoint\cite{deepjoint} &  24.20/0.8924/0.8477/0.0462 & 23.74/0.8837/0.8469/0.0483 & 22.96/0.8644/0.8366/0.0521 & 22.18/0.8432/0.8214/0.0573 \\
		Kokkinos\cite{deepunfold} &  22.51/0.8719/0.8071/0.0625 & 22.46/0.8774/0.8449/0.0533 & 21.57/0.8211/0.8303/0.0729 & 20.62/0.7880/0.8031/0.0968 \\
		ConvIR\cite{convir} &  27.23/0.9331/0.9161/0.0246 & 26.18/0.9205/0.9073/0.0268 & 24.99/0.9009/0.8933/0.0303 & 23.91/0.8831/0.8761/0.0350 \\
		FSNet\cite{fsnet} & 27.71/0.9363/0.9200/0.0220 & 26.62/0.9235/0.9096/0.0224 & 25.20/0.9026/0.8933/0.0254 & 23.87/0.8801/0.8685/0.0312 \\
		GRL\cite{grl} & {29.43}/{0.9429}/{0.9239}/{0.0204} & {28.28}/{0.9335}/{0.9237}/{0.0186} & {26.90}/{0.9190}/{0.9171}/{0.0198} & {25.68}/{0.9060}/{0.9052}/{0.0238} \\
		FourierISP\cite{fourierisp} &  26.12/0.9237/0.9067/0.0274 & 25.57/0.9170/0.9034/0.0271 & 24.67/0.8996/0.8965/0.0286 & 23.73/0.8833/0.8855/0.0319 \\
		SANet\cite{sanet} &  25.30/0.9172/0.8965/0.0364 & 24.98/0.9090/0.8950/0.0337 & 24.20/0.8925/0.8861/0.0334 & 23.33/0.8760/0.8713/0.0362 \\
		MGCC\cite{mgcc} &  {28.01}/{0.9424}/{0.9273}/{0.0268} & {27.31}/{0.9391}/{0.9232}/{0.0260} & {26.23}/{0.9306}/{0.9123}/{0.0278} & {25.24}/{0.9197}/{0.8982}/{0.0318}\\
		DFENet-L (Ours) & \textbf{32.52}/\textbf{0.9797}/\textbf{0.9691}/\textbf{0.0132} & \textbf{30.51}/\textbf{0.9698}/\textbf{0.9662}/\textbf{0.0132} & \textbf{28.71}/\textbf{0.9536}/\textbf{0.9584}/\textbf{0.0151} & \textbf{27.22}/\textbf{0.9396}/\textbf{0.9468}/\textbf{0.0198} \\
		\bottomrule

	\end{tabular}
\end{table*}

\begin{figure*}[!ht]
	\centering
	\includegraphics[width=\linewidth]{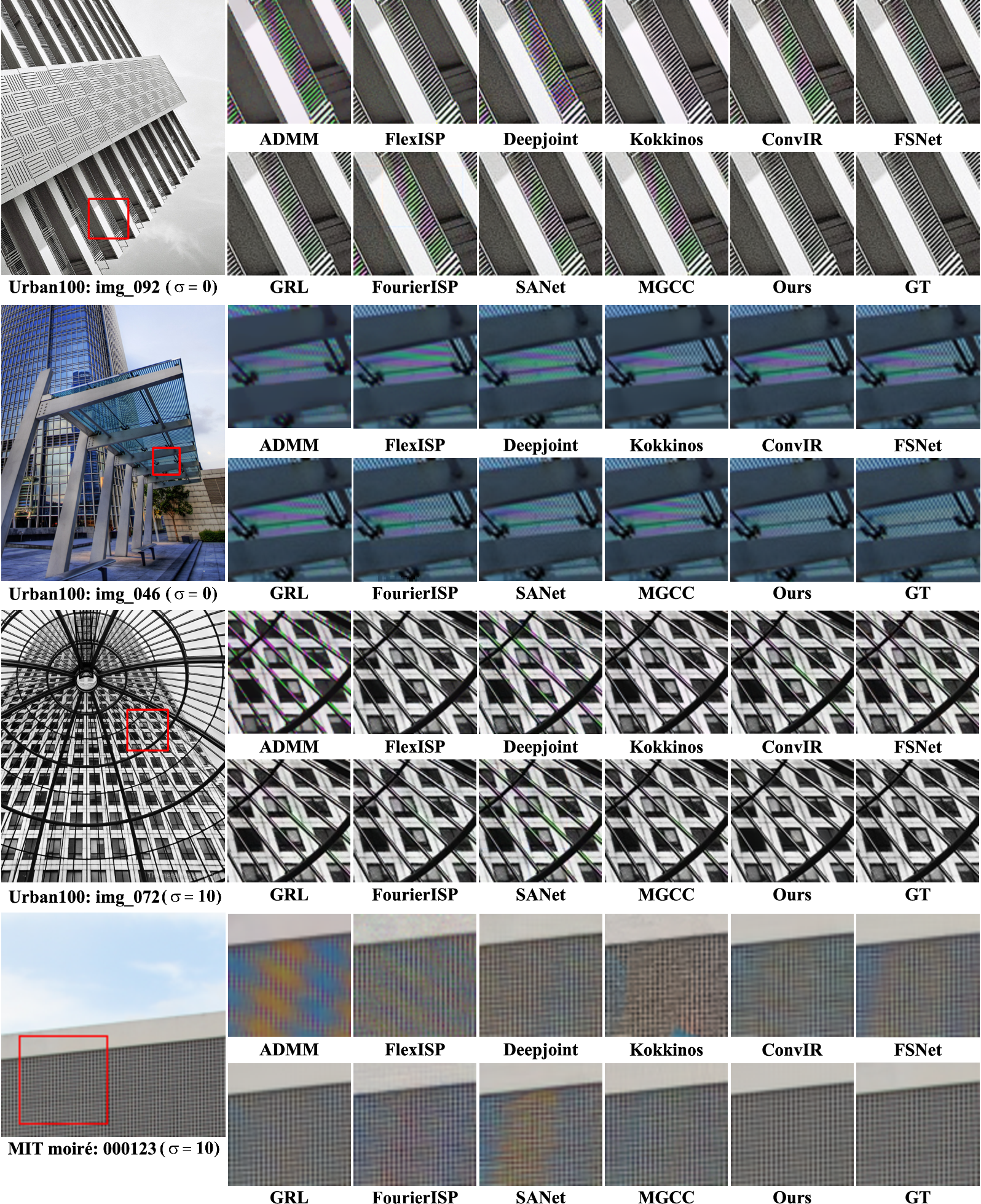}
	\caption{Visual comparison on Urban100 and MIT moir\'e among the proposed method and other compared methods.}
	\label{fig:comparison}
\end{figure*}

\begin{figure*}[!t]
	\centering
	\includegraphics[width=0.9\linewidth]{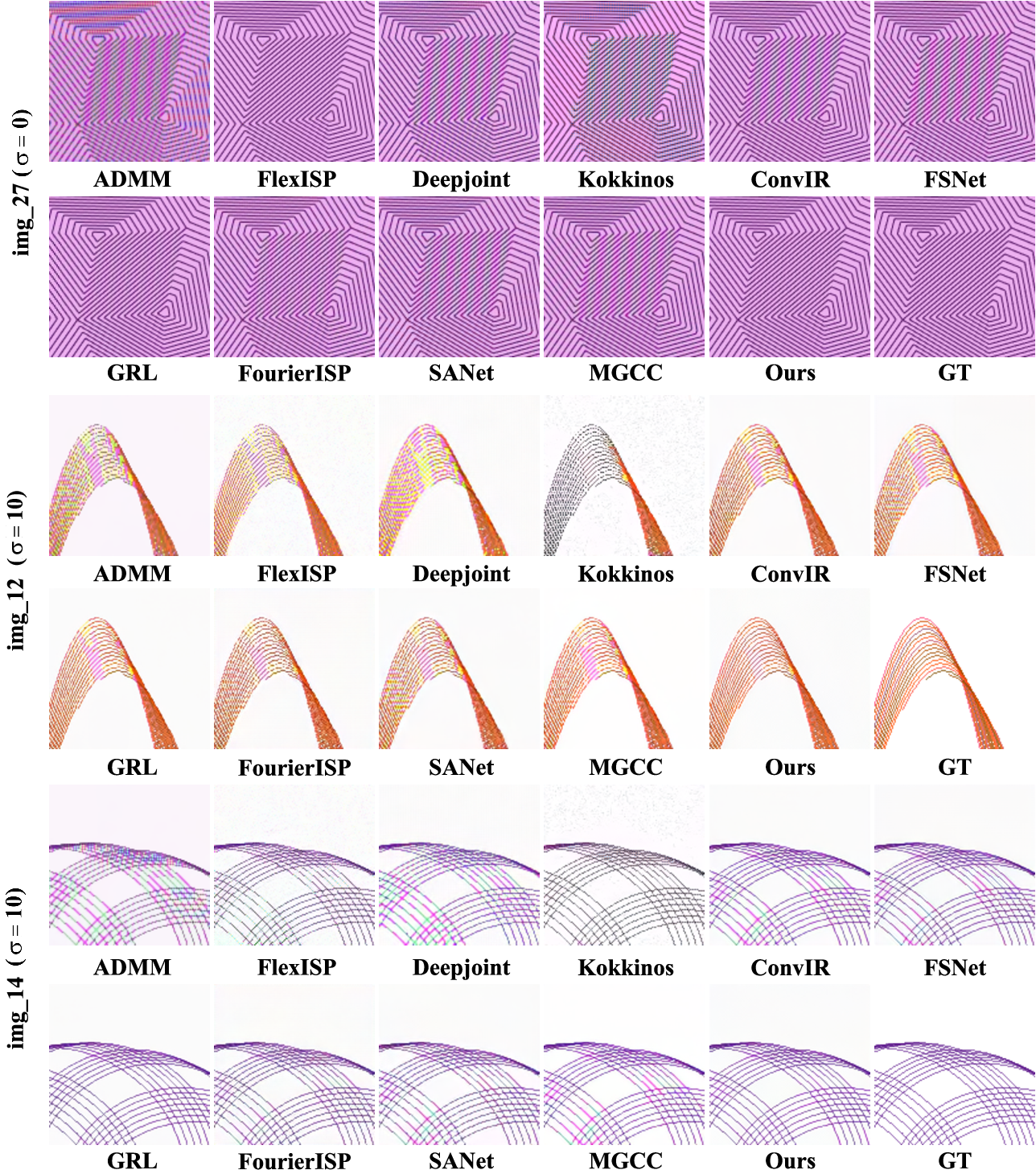}
	\caption{Visual comparison on LineSet37 among the proposed method and other compared methods. It can be observed that our reconstructions recover more accurate structures with less color moir\'e.}
	\label{fig:comparison_textures}
\end{figure*}

\section{Experiments}\label{experiments}

\subsection{Setup}

We train the model on DIV2K \cite{div2k}, which contains high-quality 800 images in the training set, and Flickr2K \cite{flickr2k} with 2650 images. For the test data, four benchmark datasets and the proposed dataset are adopted, including Kodak\footnote{https://r0k.us/graphics/kodak/}, Set14 \cite{set14}, Urban100 \cite{urban100}, MIT moir\'e \cite{deepjoint} and LineSet37.

During the training phase, RGB images are randomly cropped to $128\times 128$ and go through the augmentations by flipping and rotation. Then random Gaussian noises with variance in the range $[0, 20]$ are added to the image similar to \cite{sgnet,mgcc}. Finally, the noisy image will be sampled according to the Bayer pattern. In the testing phase, to obtain more precise fourier representations of local regions, high-resolution images will be cropped into $128\times 128$ overlapping patches, which will be reconstructed separately and combined to form the full RGB image by averaging the overlapping pixels.

The network is optimized with the Adam \cite{adam} ($\beta_1=0.9, \beta_2=0.999$) and learning rate decreasing from $10^{-4}$ to $10^{-6}$ through cosine annealing \cite{cosineannealing}. The whole training contains 500000 iterations. All experiments are implemented on Pytorch framework with NVIDIA RTX 3060 GPUs.

\subsection{Comparisons on Benchmarks}

We compare the results of ADMM \cite{admm}, FlexISP \cite{flexisp}, Deepjoint \cite{deepjoint}, Kokkinos \cite{deepunfold}, SGNet \cite{sgnet}, Wild-JDD \cite{wildjdd}, MGCC \cite{mgcc}, SANet \cite{sanet}, ConvIR \cite{convir}, FSNet \cite{fsnet}, FourierISP \cite{fourierisp} and GRL \cite{grl}. Among these methods, ConvIR and FSNet are state-of-the-art CNN-based image restoration methods, GRL is a Transformer-based image restoration method, FourierISP is a state-of-the-art ISP algorithm, and the others are demosaicking algorithms. For Deepjoint, Kokkinos and MGCC, the pre-trained model provided by the authors are used for testing. For SANet, ConvIR, FSNet and FourierISP, we retrain these models with DIV2K and Flickr2K under the same configuration and optimization strategy in their source code. The iterative times of FlexISP and ADMM are set to 30 and 50 respectively for each input image. For SGNet and Wild-JDD, we directly cite the results reported in their corresponding papers as their complete codes are not available.

The results with different noise levels $[0, 5, 10, 15]$ on benchmark datasets are given in Tab. \ref{tab:comparison}, with PSNR, SSIM \cite{ssim}, IW-SSIM \cite{iw_ssim} and LPIPS \cite{lpips} adopted for quantitative evaluation. IW-SSIM weights the scores of each pixel according to the local information content before pooling, thus it can better reflects the reconstruction quality of important local regions compared with SSIM.
From the table, it is observed that DFENet can outperform other methods by a large margin in most cases, demonstrating its effectiveness on reconstructing general natural images. Notably, for noise-free input, DFENet surpasses the second best method by approximately 0.57-1.04dB in terms of PSNR on four datasets. When noise levels increase, performance differences among different algorithms decrease.
A possible reason is that noise distorts the texture details in the input images, making it more challenging to recover the original details for all these algorithms. Moreover, the advantage of the proposed method in IW-SSIM is more pronounced compared to SSIM, highlighting its superior capability in recovering local details in regions with more information content. Additionally, FourierISP demonstrates a noticeable advantage in LPIPS on some datasets, likely due to its use of VGG loss \cite{vgg} during training. However, it still underperforms compared to other methods across most cases.


Visual comparisons are illustrated in Fig. \ref{fig:comparison}. 
While FourierISP achieves favorable LPIPS scores on Urban100 and MIT moir\'e datasets, its ability to accurately restore textures and avoid color distortion remains significantly limited. Optimization-based methods like FlexISP and Kokkinos effectively reduce color moir\'e in some images but fail to consistently deliver good results across different images. Additionally, though GRL achieves comparable quantitative scores to our method, it is notably less effective at mitigating color moir\'e.
However, the proposed DFENet consistently produces images with better visual quality compared with other methods.

\subsection{Comparisons on LineSet37}

Results are given in Table \ref{tab:comparison_textures}. As images from LineSet37 are all hard cases that are prone to severe artifacts after demosaicking, results on this dataset better reflect different algorithms' performance in reconstructing challenging structures.
It is observed that DFENet significantly outperforms other methods in terms of various metrics. Combined with the experimental results on benchmark datasets, it is demonstrated that the proposed DFENet not only achieves satisfactory demosaicking results on general natural images, but also excels in reconstructing challenging textures.

Visual results are shown in Fig. \ref{fig:comparison_textures}.
The complex line patterns in the LineSet37 pose significant challenges for demosaicking, resulting in large-scale color distortion. The proposed method outperforms others in avoiding false colors and preserving accurate texture details. Slight color discrepancies are still observed in some images when compared with the ground truth images. This is because the lines are typically only one pixel wide, and the colors of adjacent lines are not always identical. This leads to severe color information loss after sampling, making it difficult to fully recover the original colors. However, such discrepancies are acceptable, as they are much less obvious compared to color distortions produced by other methods.

\subsection{Ablation Studies}

In this section, ablation studies are performed to verify the effectiveness of the proposed method.

\noindent\textbf{Frequency Suppression.} The proposed method suppresses undesirable frequencies under the guidance of the fourier representation of the CFA images. The effectiveness of frequency suppressor in the fourier domain is explored, whose results are presented in Table \ref{tab:suppress}. The baseline model comprises residual groups with $N$ repeated RCABs only.
'W.o. suppression' is the model without frequency suppression, where frequency selection is performed and selective frequency components in the corresponding path are directly concatenated with output of the other path. There is a clear improvement over the baseline model, demonstrating the effectiveness of frequency selection. 'W.o. guidance' represents the frequency suppression performed without the guidance of the CFA images, which outperforms the results of 'w.o. suppression' in most cases. It is because frequency filtering can bring performance improvement, which has been utilized by numerous works. Nonetheless, without the guidance of the CFA images, it is challenging to detect false frequencies and further eliminate them, which explains why there is a significant performance gap compared with the results produced by 'w. guidance' on LineSet37. 
However, not all incorrect information can be reflected in the differences between the fourier representation of the CFA images and intermediate features, so 'w.o. guidance' and 'w. guidance' show similar results in some cases.

\begin{table}[!htbp]
	\centering
	\caption{Ablation studies of the frequency suppression. W.o. suppression is the model without frequency suppression. W.o. guidance and w. guidance respectively generates suppressors without and with the guidance of the CFA images.}
	\label{tab:suppress}
    \scriptsize
	\begin{tabular}{ccccc}
		\toprule
		\multirow{2}{*}{Methods} & \multirow{2}{*}{$\sigma$} & \multirow{2}{*}{Params.} & Urban100 & LineSet37 \\
		\cmidrule{4-5}
		&&& PSNR/SSIM & PSNR/SSIM \\
		
		\midrule
		baseline & \multirow{4}{*}{0} & 11.13M & 40.61/0.9861 & 31.33/0.9736 \\
		w.o. suppression &  & 11.46M  & 40.56/0.9858 & 31.76/0.9762  \\
		w.o. guidance && 11.58M & 40.85/0.9867 & 31.34/0.9749 \\
		w. guidance && 11.58M & 40.88/0.9866 & 32.04/0.9751 \\
		
		\midrule
		baseline & \multirow{4}{*}{10} & 11.13M & 33.51/0.9287 & 27.98/0.9470 \\
		w.o. suppression &  & 11.46M  & 33.49/0.9284  & 28.12/0.9474  \\
		w.o. guidance && 11.58M & 33.63/0.9196 & 28.16/0.9489 \\
		w. guidance && 11.58M & 33.66/0.9299 & 28.47/0.9485 \\
		\bottomrule
		
	\end{tabular}
\end{table}

\begin{figure}[!htbp]
	\includegraphics[width=\linewidth]{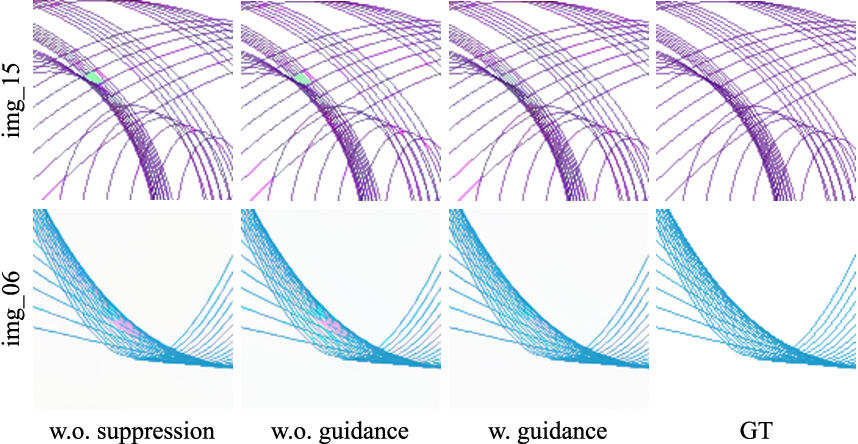}
	\caption{Visual comparison between 'w.o. suppression', 'w.o. guidance' and 'w. guidance' on LineSet37. Images on two rows are respectively produced from noiseless and noisy ($\sigma=10$) input.
		It can be observed that suppression without the guidance of the input CFA images does not benefit the alleviation of moir\'e effect, but introducing the guidance can significantly eliminate color moir\'e.}
	\label{fig:suppress_vis}
\end{figure}

\begin{figure}[!ht]
	\centering
	\includegraphics[width=\linewidth]{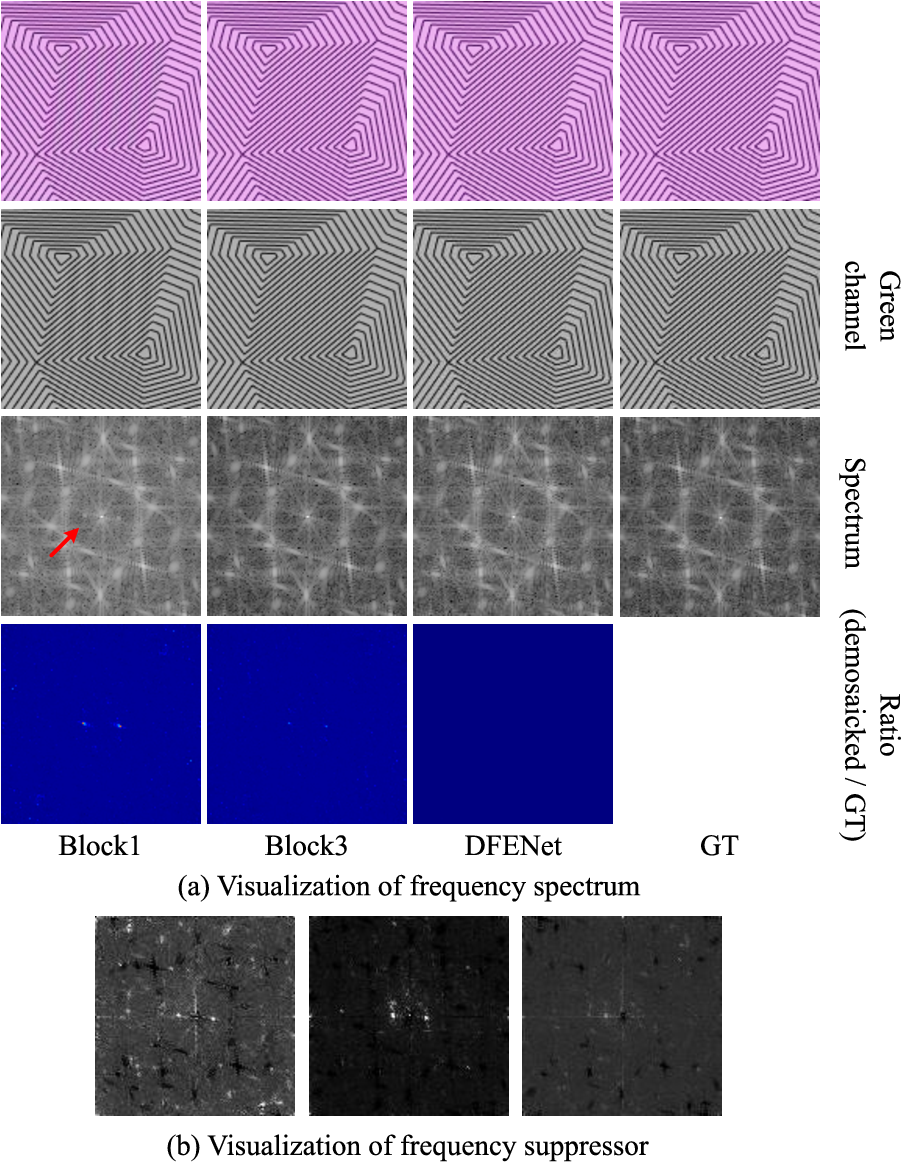}
	\caption{
		Visualization based on image (img\_27) from LineSet37.
		(a) Block1 and Block3 represent the intermediate outputs from the first and third RFEG. The first row shows the ground truth image, intermediate outputs and final output. The second row shows the green channel of these images, and the third row visualizes their frequency spectrum. To emphasize the undesirable frequency, heatmaps are given in the third row, showing the ratio between frequency of corresponding demosaicked output and the ground truth image. 
		This set of images show that the network progressively recovers accurate textures and suppressing the undesirable frequencies (two dots pointed by the arrow).
		(b) Visualization of frequency suppressors generated from the false frequency detection module. The generated masks can effectively identify undesirable frequencies and distinguish them with correct frequencies that should be preserved.}
	\label{fig:suppress}
\end{figure}

To show the importance of the guidance from CFA image, demosaicked results from 'w.o. suppression', 'w.o. guidance' and 'w. guidance' are given in Fig. \ref{fig:suppress_vis}. Images from the two rows are respectively produced from noisy-free and noisy ($\sigma=10$) input. Notably, frequency suppression without the guidance of input CFA images fails to bring performance improvement, where obvious color distortion still exist. However, after introducing the guidance, color moir\'e can be effectively alleviated as shown in the results from 'w. guidance'.

To further show the efficacy of the proposed frequency suppression path, analysis based on generated frequency suppressor is given and corresponding visualization is illustrated in Fig. \ref{fig:suppress}. 
Comparing the intermediate output from the first RFEG (Block1) and third RFEG (Block3) with the final demosaicked result of DFENet, it can be observed that the network progressively recovers the textures. Shallower layers produce image with obvious moir\'e pattern and such false textures are diminished in deeper layers. Accordingly, as shown in Fig. \ref{fig:suppress} (a), undesirable low frequencies (two dots pointed by the arrow) produced in the shallow layers are also progressively suppressed and eliminated. 
FFD plays an important role in achieving this. 
We visualize the frequency suppressors generated by FFD from different layers, as shown in Fig. \ref{fig:suppress} (b). To better showcase the capability of this module and facilitate the observation of visualized results, the visualization is given based on the fourier representations without performing frequency selection. Since only $1\times 1$ convolution is used in FFD, frequency selection does not have impact on the generation of suppressors. It can be observed that the network can effectively identify undesirable frequencies and distinguish them with those correct frequencies that should be preserved.
Therefore, it can be concluded that the frequency suppression path can achieve its intended effect. However, it is impossible to completely eliminate those false frequencies because they may be passed to the deeper layer through skip connections or the other path of the SFE. In future work, we can further explore more stringent methods to block the propagation of erroneous information.

\noindent\textbf{Frequency Selection}. The proposed method introduces two learnable frequency selectors to decide which frequencies should be concentrated on missing information generation and which on false frequency suppression. Results that 'w.o. suppression' outperforms the baseline model have demonstrated the effectiveness of frequency selection. 
Experiments in this part explores different selection methods, whose performance are given in Table \ref{tab:selection}. FSNet performs frequency selection in spatial domain with dynamic convolutional kernels and its inverted kernels. 
'Fixed' represents a fixed threshold that is manually set to divide all frequencies into high and low frequencies, where frequencies inside the circle around the DC component with a radius of 0.5$\pi$ are low frequencies. 
'Single selector' uses a single learnable selector to choose frequencies to go through the path that focuses on spatial-domain detail refinement and the rest frequencies to go through another path, while 'dual selector' uses two learnable selectors for the two paths.

\begin{table}[!t]
	\centering
	\caption{Results of different frequency selection methods.}
	\label{tab:selection}
    \scriptsize
	\begin{tabular}{ccccc}
            
		\toprule
		\multirow{2}{*}{Methods} &\multirow{2}{*}{ $\sigma$} & \multirow{2}{*}{Params.} & MIT moir\'e  & LineSet37\\
		\cmidrule{4-5}
		&&& PSNR/SSIM & PSNR/SSIM\\
		\midrule
		FSNet\cite{fsnet} & \multirow{5}{*}{0} & 11.60M  & 36.88/0.9429 & 30.72/0.9719 \\
		fixed && 11.24M  & 37.50/0.9455 & 31.15/0.9722\\
		single selector && 11.24M  & 37.49/0.9453 & 31.46/0.9746\\
		dual selector && 11.58M  &37.61/0.9457 & 32.04/0.9751 \\
		\midrule
		FSNet\cite{fsnet} & \multirow{5}{*}{10} & 11.60M  & 31.50/0.8830 & 27.87/0.9466 \\
		fixed && 11.24M  & 31.71/0.8853 & 28.07/0.9479\\
		single selector && 11.24M  & 31.72/0.8863 & 28.08/0.9463\\
		dual selector && 11.58M  & 31.85/0.8889 & 28.47/0.9485\\
		\bottomrule
	\end{tabular}
\end{table}

From Table \ref{tab:selection}, it is observed that selection in spatial domain fails to effectively decide which features lack high-frequency information and which contain undesirable information. 
Fixed selectors with a manually set threshold achieve satisfactory performance though the fourier representations vary across different images. 
Learnable 'single selector' substantially surpasses 'fixed' on LineSet37 when processing noise-free images. However, binary partitions face a disadvantage when some components contain both missing frequencies and undesirable frequencies, limiting its performance on some images.
By using learnable selectors for both paths, the best results can be achieved. 

\noindent\textbf{Exploring resolution of frequency selectors.} As directly generating selectors with the same size as the image will lead to overly fragmented information, low-resolution selectors are learned and upscaled to the target resolution before performing frequency selection. In Table \ref{tab:downfactor}, the resolution of frequency selectors is explored. Since the full images will be cropped to $128\times 128$ before feeding into the network, the max size of frequency selectors is set to $128\times 128$. Other sizes $64\times 64$, $32\times 32$, $16\times 16$ and $8\times 8$ are given in the table. 
Performance improvement can be observed when reducing the size of frequency selectors, especially on LineSet37 and Urban100. However, after decreasing the size from $16\times 16$ to $8\times 8$, results on noisy images in LineSet37 show a noticeable performance drop. It is because two small selectors lead to too coarse selection, limiting model's performance.

\begin{table}[!t]
	\caption{Different sizes of frequency selectors. Best results are highlighted in \textbf{bold}.}
	\label{tab:downfactor}
	\centering
    \scriptsize
	\begin{tabular}{ccccc}
    
		\toprule
		\multirow{2}{*}{Size of Selectors} & \multirow{2}{*}{$\sigma$} &  Urban100 & MIT moir\'e & LineSet37\\
		\cmidrule{3-5}
		&& PSNR/SSIM & PSNR/SSIM & PSNR/SSIM\\
		\midrule
		128 & \multirow{5}{*}{0}  & 40.85/0.9864 & 37.65/0.9462 & 31.71/0.9720 \\
		64 & & 40.75/0.9863 & 37.63/0.9462 & 31.55/0.9762 \\
		32 & & 40.80/0.9865 & 37.65/0.9464 & 32.00/0.9774 \\
		16 & & 40.88/\textbf{0.9866} & 37.61/0.9457 & \textbf{32.04}/0.9751 \\
		8 && \textbf{40.90}/0.9865 & \textbf{37.66}/\textbf{0.9465} & 32.03/\textbf{0.9777}  \\
		\midrule
		128 & \multirow{5}{*}{10} & 33.63/0.9294 & 31.85/0.8882 & 28.23/0.9479  \\
		64 & & 33.62/0.9295 & 31.84/0.8881 &  28.31/0.9499 \\
		32 && 33.63/\textbf{0.9358} & 31.86/0.8889 &  28.37/\textbf{0.9505} \\
		16 && \textbf{33.66}/0.9299 & 31.85/\textbf{0.8889} &  \textbf{28.47}/0.9485 \\
		8 && 33.65/0.9297 & \textbf{31.86}/0.8888 & 28.28/0.9501  \\
		\bottomrule
	\end{tabular}
\end{table}

\noindent\textbf{Number of RCABs for selected frequencies.} The proposed DFENet enhances the efficiency of the network by allowing part of RCABs to focus on selected features, unlike previous methods where all frequencies go through the same number of blocks. Under the same total number of RCABs, we explore the impact of the ratio between $N_1$ and $N_2$ on the demosaicking performance.  
Table \ref{tab:numblock} shows the results with different ratios between $N_2$ and total number of RCABs in each RFEG. Except for the configurations with the largest ratio, the results on the validation set are similar for other ratios. The configuration $N_1=N_2$ is used for the proposed method. 

\begin{table}[!ht]
	\centering
	\caption{Quantitative results (PSNR/SSIM) on DIV2K validation dataset. $N_1$ and $N_2$ are respectively the number of blocks for coarse feature extraction and detail refinement.}
	\label{tab:numblock}
    \scriptsize
	\begin{tabular}{ccc}
    
			\toprule
			$N_1$:$N_2$ &  $\sigma=0$ &  $\sigma=10$ \\
			\midrule
			3:1 & 43.5766/0.9907 & 34.6897/0.9181 \\
			2:2 & 43.5818/0.9907 & 34.6920/0.9181 \\
			1:3 & 43.5770/0.9907 & 34.6921/0.9180 \\
			0:4 & 43.5094/0.9906 & 34.6799/0.9182 \\
			\bottomrule
		\end{tabular}
\end{table}

\noindent\textbf{FFT vs. DCT.} Discrete cosine transformation (DCT) has been utilized by numerous restoration tasks \cite{freqmoire} and achieved excellent performance, with some outperforming FFT-based processings. However, though DCT offers advantages such as energy compaction and computational efficiency, FFT has the advantage of providing more comprehensive frequency information, making it better suited for frequency analysis. Table \ref{tab:dct} reveals the results of the model where FFT is replaced by DCT, including frequency selection, frequency suppression and frequency supervision. FFT demonstrates significantly better performance compared to DCT.

\begin{table}[!thbp]
	\centering
	\caption{Comparison between DCT and FFT. FFT is replaced in frequency selection, frequency suppression and frequency supervision.}
	\label{tab:dct}
    \scriptsize
	\begin{tabular}{ccccc}
		\toprule
		\multirow{2}{*}{Transformation} & \multirow{2}{*}{$\sigma$} & \multirow{2}{*}{Params.} & MIT moir\'e & LineSet37 \\
		\cmidrule{4-5}
		&&& PSNR/SSIM & PSNR/SSIM\\
		\midrule
		FFT & \multirow{2}{*}{0} & 11.58M & 37.64/0.9453 & 32.92/0.9813 \\
		DCT && 11.55M & 37.55/0.9455 & 32.83/0.9793 \\
		\midrule
		FFT & \multirow{2}{*}{10} & 11.58M &31.81/0.8875 & 28.75/0.9538  \\
		DCT && 11.55M & 31.75/0.8865 & 28.38/0.9516 \\
		\bottomrule
	\end{tabular}
\end{table}

\noindent\textbf{Optimization.} In Table \ref{tab:loss}, multi-level frequency supervision and a stagewise training strategy is used. Without suitable optimization, the effects of frequency loss are limited on some datasets. With a stagewise training strategy where different supervisions are used for two separate backpropagation processes to optimize the network independently, considerable performance improvement can be achieved.

\begin{table}[!t]
\caption{The results of introduction of multi-level frequency supervision and stagewise training strategy. }
\label{tab:loss}
    \centering
    \scriptsize
    \begin{tabular}{ccccc}
    \toprule
    \multirow{2}{*}{Optimization} & \multirow{2}{*}{$\sigma$} & Kodak  & Urban100 & MIT moir\'e \\
    \cmidrule{3-5}
    && PSNR/SSIM & PSNR/SSIM & PSNR/SSIM\\
    \midrule
    L1 loss & \multirow{3}{*}{0} & 43.38/0.9905 & 40.88/0.9866 & 37.61/0.9457 \\
    + FFT loss && 43.37/0.9905 & 41.06/0.9869 & 37.64/0.9453 \\
    + Stagewise && 43.39/0.9905 & 41.25/0.9871 & 37.82/0.9461  \\
    \midrule
    L1 loss & \multirow{3}{*}{10} & 34.32/0.9181 & 33.66/0.9299 & 31.85/0.8889\\
    + FFT loss && 34.29/0.9178 & 33.63/0.9298 & 31.81/0.8875  \\
    + Stagewise && 34.31/0.9179 & 33.67/0.9300 & 31.89/0.8890  \\
    \bottomrule
    \end{tabular}
\end{table}

\begin{table}[t]
	\centering
	\caption{Results with and without the TLC.}
	\label{tab:tlc}
    \scriptsize
	\begin{tabular}{ccccc}
		\toprule
		\multirow{2}{*}{Methods} & \multirow{2}{*}{$\sigma$} & Kodak & Urban100 & MIT moir\'e \\
		\cmidrule{3-5}
		&& PSNR/SSIM & PSNR/SSIM & PSNR/SSIM \\
		\midrule
		baseline & \multirow{4}{*}{0} & 43.33/0.9905 & 40.64/0.9861 & 37.22/0.9445  \\
		baseline+TLC && 43.33/0.9905 & 40.61/0.9861 & 37.22/0.9445  \\
		DFENet && 42.54/0.9889 & 39.17/0.9818 & 37.82/0.9461 \\
		DFENet+TLC && 43.39/0.9905 & 41.25/0.9871 & 37.82/0.9461  \\
		\midrule
		baseline & \multirow{4}{*}{10} & 34.28/0.9174 & 33.52/0.9288 & 31.47/0.8822 \\
		baseline+TLC && 34.27/0.9174 & 33.51/0.9287 & 31.47/0.8822  \\
		DFENet && 30.35/0.7783 & 29.69/0.8135 & 31.89/0.8890  \\
		DFENet+TLC && 34.31/0.9179 & 33.67/0.9300 & 31.89/0.8890  \\
		\bottomrule
	\end{tabular}
\end{table}

\noindent\textbf{Use of the Test-time Local Converter.} As the model is trained on $128\times 128$ patches while the size of the test images can be much larger, distribution gap exists between the training and testing phases.
Such distribution gap significantly restricts the performance of DFENet. It is because fourier transformation, which encodes the global information of the entire images, makes false frequencies caused by local artifacts like color moir\'e easily overlooked. To address this, test-time Local Converter (TLC) \cite{tlc} is introduced to reduce the inconsistency of the two phases, which separately processes the cropped overlapping patches and integrates the overlapping regions by averaging. Table \ref{tab:tlc} reveals the results with and without TLC. For the baseline without operations in the fourier domain, consistent results are obtained with or without the TLC.
For the proposed model, without the TLC, performance on MIT moir\'e with $128\times 128$ images remains the same, but significant performance decline occurs on Urban100 and Kodak, yielding poorer results than the baseline. This indicates that the fourier operations are sensitive to the image sizes. With the TLC, the problem can be avoided.

\subsection{Model Complexity} 

Table \ref{tab:complexity} compares the model complexity and performance of different algorithms. DFENet-S and DFENet-T are small and tiny versions of the proposed method. It can be observed that our method consistently outperforms the comparison methods with equivalent amount of parameters across different model complexities, demonstrating the superiority of our method.

\begin{table}[!htbp]
	\centering
	\caption{Comparison of model complexity and performance on LineSet37.}
	\label{tab:complexity}
    \scriptsize
	\begin{tabular}{cccc}
		\toprule
		\multirow{2}{*}{Methods} & \multirow{2}{*}{Params.} & $\sigma=0$& $\sigma=10$ \\
		\cmidrule{3-4}
		&& PSNR/SSIM & PSNR/SSIM\\
		\midrule
		
		ADMM\cite{admm} & - & 18.89/0.7595 & 19.22/0.7581   \\
		FlexISP\cite{flexisp} & - & 25.79/0.8976 & 23.22/0.8459  \\
		Deepjoint\cite{deepjoint} & 0.56M & 24.20/0.8924 & 22.96/0.8644 \\
		Kokkinos\cite{deepunfold} & 0.38M & 22.51/0.8719 & 21.57/0.8211 \\
		DFENet-T (Ours) & 0.12M & \textbf{26.44}/\textbf{0.9393} & \textbf{25.08}/\textbf{0.9186} \\
		\hdashline
		
		ConvIR\cite{convir} & 5.53M & 27.23/0.9331& 24.99/0.9009 \\
		FSNet\cite{fsnet} & 3.95M & 27.71/0.9363  & 25.20/0.9026  \\
		GRL\cite{grl} & 3.34M & {29.43}/{0.9429}  & {26.90}/{0.9190} \\
		DFENet-S (Ours) & 3.37M & \textbf{31.10}/\textbf{{0.9703}} & \textbf{27.86}/\textbf{0.9437} \\
		\hdashline

		FourierISP\cite{fourierisp} & 7.59M & 26.12/0.9237  & 24.67/0.8996 \\
		SANet\cite{sanet} & 10.78M & 25.30/0.9172  & 24.20/0.8925\\
		MGCC\cite{mgcc} & 13.78M & {28.01}/{0.9424} & {26.23}/{0.9306} \\
		
		DFENet (Ours) & 11.58M & \textbf{32.52}/\textbf{0.9797} & \textbf{28.71}/\textbf{0.9536} \\
		\bottomrule

	\end{tabular}
\end{table}

\section{Conclusion}\label{conclusion}

The paper proposed an image demosaicking method with enhanced ability of reconstructing challenging structures. Based on frequency analysis of results produced from benchmark algorithms, by introducing learnable frequency selectors, different frequencies are processed differently. Some frequencies focus on generating missing information while others focus on suppressing false frequency. Experiments demonstrate that the proposed method outperforms other state-of-the-art methods both quantitatively and qualitatively. Extensive ablation studies also prove the efficacy of the designs. Moreover, to better evaluate different algorithms' performance on hard cases, a new dataset, LineSet37, is contributed, with 37 artificially designed and generated images featuring complex line patterns. Score differences on this dataset can provide a more targeted assessment of algorithms' effectiveness in reconstructing difficult structures.

\bibliography{main}

\begin{thebibliography}{46}
\providecommand{\natexlab}[1]{#1}
\providecommand{\url}[1]{\texttt{#1}}
\expandafter\ifx\csname urlstyle\endcsname\relax
  \providecommand{\doi}[1]{doi: #1}\else
  \providecommand{\doi}{doi: \begingroup \urlstyle{rm}\Url}\fi

\bibitem[Bayer(1976)]{bayer}
Bryce Bayer.
\newblock Color imaging array, Jul. 1976.

\bibitem[Pei and Tam(2000)]{colordifference}
Soo-Chang Pei and Io-Kuong Tam.
\newblock Effective color interpolation in {CCD} color filter array using signal correlation.
\newblock In \emph{Proc. {IEEE} Int. Conf. Image Process. (ICIP)}, volume~3, pages 488--491, Sept. 2000.
\newblock \doi{10.1109/ICIP.2000.899455}.

\bibitem[Hirakawa and Parks(2005)]{hirakawa}
K.~Hirakawa and T.W. Parks.
\newblock Adaptive homogeneity-directed demosaicing algorithm.
\newblock \emph{{IEEE} Trans. Image Process.}, 14\penalty0 (3):\penalty0 360--369, Feb. 2005.
\newblock \doi{10.1109/TIP.2004.838691}.

\bibitem[Li(2005)]{successive}
Xin Li.
\newblock Demosaicing by successive approximation.
\newblock \emph{{IEEE} Trans. Image Process.}, 14\penalty0 (3):\penalty0 370--379, Mar. 2005.

\bibitem[Zhang and Wu(2005)]{lmmse}
Lei Zhang and Xiaolin Wu.
\newblock Color demosaicking via directional linear minimum mean square-error estimation.
\newblock \emph{{IEEE} Trans. Image Process.}, 14\penalty0 (12):\penalty0 2167--2178, Sept. 2005.
\newblock \doi{10.1109/TIP.2005.857260}.

\bibitem[Chen et~al.(2015)Chen, He, Jeon, and Jeong]{multidirectional}
Xiangdong Chen, Liwen He, Gwanggil Jeon, and Jechang Jeong.
\newblock Multidirectional weighted interpolation and refinement method for bayer pattern {CFA} demosaicking.
\newblock \emph{{IEEE} Trans. Circuits Syst. Video Technol.}, 25\penalty0 (8):\penalty0 1271--1282, Mar. 2015.
\newblock \doi{10.1109/TCSVT.2014.2313896}.

\bibitem[Wang and Jeon(2015)]{multi2}
Lei Wang and Gwanggil Jeon.
\newblock Bayer pattern {CFA} demosaicking based on multi-directional weighted interpolation and guided filter.
\newblock \emph{{IEEE} Signal Process. Lett.}, 22\penalty0 (11):\penalty0 2083--2087, Jul. 2015.

\bibitem[Buades et~al.(2009)Buades, Coll, Morel, and Sbert]{buades}
Antoni Buades, Bartomeu Coll, Jean-Michel Morel, and Catalina Sbert.
\newblock Self-similarity driven color demosaicking.
\newblock \emph{{IEEE} Trans. Image Process.}, 18\penalty0 (6):\penalty0 1192--1202, Apr. 2009.
\newblock \doi{10.1109/TIP.2009.2017171}.

\bibitem[Duran and Buades(2014)]{duran}
Joan Duran and Antoni Buades.
\newblock Self-similarity and spectral correlation adaptive algorithm for color demosaicking.
\newblock \emph{{IEEE} Trans. Image Process.}, 23\penalty0 (9):\penalty0 4031--4040, Jul. 2014.

\bibitem[Alleysson et~al.(2002)Alleysson, Süsstrunk, and Herault]{fft1}
David Alleysson, Sabine Süsstrunk, and Jeanny Herault.
\newblock Color demosaicing by estimating luminance and opponent chromatic signals in the fourier domain.
\newblock In \emph{Proc. Color Imaging Conf.}, volume~10, Nov. 2002.
\newblock \doi{10.2352/CIC.2002.10.1.art00061}.

\bibitem[Alleysson et~al.(2005)Alleysson, Susstrunk, and Herault]{fft2}
D.~Alleysson, S.~Susstrunk, and J.~Herault.
\newblock Linear demosaicing inspired by the human visual system.
\newblock \emph{{IEEE} Trans. Image Process.}, 14\penalty0 (4):\penalty0 439--449, Mar. 2005.
\newblock \doi{10.1109/TIP.2004.841200}.

\bibitem[Dubois(2005)]{fft3}
E.~Dubois.
\newblock Frequency-domain methods for demosaicking of bayer-sampled color images.
\newblock \emph{{IEEE} Signal Process. Lett.}, 12\penalty0 (12):\penalty0 847--850, Nov. 2005.
\newblock \doi{10.1109/LSP.2005.859503}.

\bibitem[Tan et~al.(2017)Tan, Zeng, Lai, Liu, and Zhang]{admm}
Hanlin Tan, Xiangrong Zeng, Shiming Lai, Yu~Liu, and Maojun Zhang.
\newblock Joint demosaicing and denoising of noisy bayer images with {ADMM}.
\newblock In \emph{Proc. {IEEE} Int. Conf. Image Process. (ICIP)}, pages 2951--2955, Sept. 2017.
\newblock \doi{10.1109/ICIP.2017.8296823}.

\bibitem[Pulli(2014)]{flexisp}
Kari Pulli.
\newblock Flexisp: a flexible camera image processing framework.
\newblock \emph{{ACM} Trans. Graph.}, 33\penalty0 (6):\penalty0 231:1--231:13, Nov. 2014.
\newblock \doi{10.1145/2661229.2661260}.

\bibitem[Gharbi et~al.(2016)Gharbi, Chaurasia, Paris, and Durand]{deepjoint}
Micha\"{e}l Gharbi, Gaurav Chaurasia, Sylvain Paris, and Fr\'{e}do Durand.
\newblock Deep joint demosaicking and denoising.
\newblock \emph{ACM Trans. Graph.}, 35\penalty0 (6), Dec. 2016.
\newblock ISSN 0730-0301.
\newblock \doi{10.1145/2980179.2982399}.

\bibitem[Liu et~al.(2020)Liu, Jia, Liu, and Tian]{sgnet}
Lin Liu, Xu~Jia, Jianzhuang Liu, and Qi~Tian.
\newblock Joint demosaicing and denoising with self guidance.
\newblock In \emph{Pro. {IEEE/CVF} Conf. Comput. Vis. Pattern Recognit. (CVPR)}, pages 2237--2246, Jun. 2020.
\newblock \doi{10.1109/CVPR42600.2020.00231}.

\bibitem[Chen et~al.(2021)Chen, Wen, and Chan]{wildjdd}
Jierun Chen, Song Wen, and S.{-}H.~Gary Chan.
\newblock Joint demosaicking and denoising in the wild: The case of training under ground truth uncertainty.
\newblock In \emph{Proc. {AAAI} Conf. Artif. Intell. (AAAI)}, pages 1018--1026, Feb. 2021.

\bibitem[Zhang et~al.(2022{\natexlab{a}})Zhang, Sun, and Chen]{mgcc}
Yong Zhang, Wanjie Sun, and Zhenzhong Chen.
\newblock Joint image demosaicking and denoising with mutual guidance of color channels.
\newblock \emph{Signal Process.}, 200:\penalty0 108674, Nov. 2022{\natexlab{a}}.
\newblock ISSN 0165-1684.
\newblock \doi{https://doi.org/10.1016/j.sigpro.2022.108674}.

\bibitem[Zhang et~al.(2022{\natexlab{b}})Zhang, Fu, and Li]{sanet}
Tao Zhang, Ying Fu, and Cheng Li.
\newblock Deep spatial adaptive network for real image demosaicing.
\newblock In \emph{Proc. {AAAI} Conf. Artif. Intell. (AAAI)}, volume~36, pages 3326--3334, Feb. 2022{\natexlab{b}}.
\newblock \doi{10.1609/aaai.v36i3.20242}.

\bibitem[Shannon(1949)]{nyquist}
Claude~Elwood Shannon.
\newblock Communication in the presence of noise.
\newblock \emph{Proc. {IRE}}, 37\penalty0 (1):\penalty0 10--21, Sept. 1949.

\bibitem[Buades et~al.(2005)Buades, Coll, and Morel]{nonlocal}
Antoni Buades, Bartomeu Coll, and Jean{-}Michel Morel.
\newblock A review of image denoising algorithms, with a new one.
\newblock \emph{{SIAM} Multiscale Model. Simul.}, 4\penalty0 (2):\penalty0 490--530, 2005.

\bibitem[Guan et~al.(2022)Guan, Lai, Lu, Li, Li, Feng, Yang, and Gu]{deformable}
Juntao Guan, Rui Lai, Yang Lu, Yangang Li, Huanan Li, Lichen Feng, Yintang Yang, and Lin Gu.
\newblock Memory-efficient deformable convolution based joint denoising and demosaicing for {UHD} images.
\newblock \emph{{IEEE} Trans. Circuits Syst. Video Technol.}, 32\penalty0 (11):\penalty0 7346--7358, Jun. 2022.

\bibitem[Elgendy et~al.(2021)Elgendy, Gnanasambandam, Chan, and Ma]{quad2}
Omar~A. Elgendy, Abhiram Gnanasambandam, Stanley~H. Chan, and Jiaju Ma.
\newblock Low-light demosaicking and denoising for small pixels using learned frequency selection.
\newblock \emph{{IEEE} Trans. Comput. Imag.}, 7:\penalty0 137--150, 2021.
\newblock \doi{10.1109/TCI.2021.3052694}.

\bibitem[Xing and Egiazarian(2021)]{xing}
Wenzhu Xing and Karen~O. Egiazarian.
\newblock End-to-end learning for joint image demosaicing, denoising and super-resolution.
\newblock In \emph{Proc. {IEEE} Conf. Comput. Vis. Pattern Recognit. (CVPR)}, pages 3507--3516, Jun. 2021.
\newblock \doi{10.1109/CVPR46437.2021.00351}.

\bibitem[Bai et~al.(2024)Bai, Li, and Wang]{freq2}
Chenyan Bai, Jia Li, and Jinbiao Wang.
\newblock Spatial-frequency fusion for bayer demosaicking.
\newblock \emph{{IEEE} Signal Process. Lett.}, 31:\penalty0 2245--2249, Aug. 2024.

\bibitem[Ma et~al.(2022)Ma, Gharbi, Adams, Kamil, Li, Barnes, and Ragan{-}Kelley]{search}
Karima Ma, Micha{\"{e}}l Gharbi, Andrew Adams, Shoaib Kamil, Tzu{-}Mao Li, Connelly Barnes, and Jonathan Ragan{-}Kelley.
\newblock Searching for fast demosaicking algorithms.
\newblock \emph{{ACM} Trans. Graph.}, 41\penalty0 (5):\penalty0 172:1--172:18, May. 2022.

\bibitem[Lee et~al.(2023)Lee, Park, Jeong, Kim, Je, Ryu, and Chun]{quad4}
Haechang Lee, Dongwon Park, Wongi Jeong, Kijeong Kim, Hyunwoo Je, Dongil Ryu, and Se~Young Chun.
\newblock Efficient unified demosaicing for bayer and non-bayer patterned image sensors.
\newblock In \emph{Proc. {IEEE/CVF} Int. Conf. Comput. Vis. (ICCV)}, pages 12704--12713, Oct. 2023.
\newblock \doi{10.1109/ICCV51070.2023.01171}.

\bibitem[Zhang et~al.(2018{\natexlab{a}})Zhang, Li, Li, Wang, Zhong, and Fu]{rcan}
Yulun Zhang, Kunpeng Li, Kai Li, Lichen Wang, Bineng Zhong, and Yun Fu.
\newblock Image super-resolution using very deep residual channel attention networks.
\newblock In \emph{Proc. Eur. Conf. Comput. Vis. (ECCV)}, volume 11211, pages 294--310, Sept. 2018{\natexlab{a}}.
\newblock \doi{10.1007/978-3-030-01234-2\_18}.

\bibitem[Li et~al.(2021)Li, Hu, Wang, Li, She, Zhu, Zhang, and Chen]{involution}
Duo Li, Jie Hu, Changhu Wang, Xiangtai Li, Qi~She, Lei Zhu, Tong Zhang, and Qifeng Chen.
\newblock Involution: Inverting the inherence of convolution for visual recognition.
\newblock In \emph{Proc. {IEEE/CVF} Conf. Comput. Vis. Pattern Recognit. (CVPR)}, pages 12321--12330, Jun. 2021.

\bibitem[Kokkinos and Lefkimmiatis(2018)]{deepunfold}
Filippos Kokkinos and Stamatios Lefkimmiatis.
\newblock Deep image demosaicking using a cascade of convolutional residual denoising networks.
\newblock In \emph{Proc. Eur. Conf. Comput. Vis. (ECCV)}, volume 11218, pages 317--333, Sept. 2018.
\newblock \doi{10.1007/978-3-030-01264-9\_19}.

\bibitem[{Y. Cui, W. Ren, X. Cao, and A. Knoll}(2024)]{convir}
{Y. Cui, W. Ren, X. Cao, and A. Knoll}.
\newblock Revitalizing convolutional network for image restoration.
\newblock \emph{{IEEE} Trans. Pattern Anal. Mach. Intell.}, pages 1--16, 2024.

\bibitem[{Y. Cui, W. Ren, X. Cao and A. Knoll}(2024)]{fsnet}
{Y. Cui, W. Ren, X. Cao and A. Knoll}.
\newblock Image restoration via frequency selection.
\newblock \emph{{IEEE} Trans. Pattern Anal. Mach. Intell.}, 46\penalty0 (2):\penalty0 1093--1108, Nov. 2024.
\newblock \doi{10.1109/TPAMI.2023.3330416}.

\bibitem[Li et~al.(2023)Li, Fan, Xiang, Demandolx, Ranjan, Timofte, and Gool]{grl}
Yawei Li, Yuchen Fan, Xiaoyu Xiang, Denis Demandolx, Rakesh Ranjan, Radu Timofte, and Luc~Van Gool.
\newblock Efficient and explicit modelling of image hierarchies for image restoration.
\newblock In \emph{Proc. {IEEE/CVF} Conf. Comput. Vis. Pattern Recognit. (CVPR)}, pages 18278--18289, Jun. 2023.
\newblock \doi{10.1109/CVPR52729.2023.01753}.

\bibitem[He et~al.(2024)He, Hu, Wang, Wang, Wang, Zhang, Yan, Chen, Li, Xie, Zhang, and Zhou]{fourierisp}
Xuanhua He, Tao Hu, Guoli Wang, Zejin Wang, Run Wang, Qian Zhang, Keyu Yan, Ziyi Chen, Rui Li, Chengjun Xie, Jie Zhang, and Man Zhou.
\newblock Enhancing raw-to-s{RGB} with decoupled style structure in fourier domain.
\newblock In \emph{Proc. {AAAI} Conf. Artif. Intell. (AAAI)}, volume~38, pages 2130--2138, Mar. 2024.

\bibitem[Agustsson and Timofte(2017)]{div2k}
Eirikur Agustsson and Radu Timofte.
\newblock {NTIRE} 2017 challenge on single image super-resolution: Dataset and study.
\newblock In \emph{Proc. {IEEE} Conf. Comput. Vis. Pattern Recognit. Workshops (CVPRW)}, pages 126--135, Jul. 2017.
\newblock \doi{10.1109/CVPRW.2017.150}.

\bibitem[Lim et~al.(2017)Lim, Son, Kim, Nah, and Lee]{flickr2k}
Bee Lim, Sanghyun Son, Heewon Kim, Seungjun Nah, and Kyoung~Mu Lee.
\newblock Enhanced deep residual networks for single image super-resolution.
\newblock In \emph{Proc. {IEEE} Conf. Comput. Vis. Pattern Recognit. Workshops (CVPRW)}, pages 1132--1140, Jul. 2017.
\newblock \doi{10.1109/CVPRW.2017.151}.

\bibitem[Zeyde et~al.(2012)Zeyde, Elad, and Protter]{set14}
Roman Zeyde, Michael Elad, and Matan Protter.
\newblock On single image scale-up using sparse-representations.
\newblock In \emph{Proc. Int. Conf. Curves Surf.}, pages 711--730, Jun. 2012.
\newblock \doi{10.1007/978-3-642-27413-8_47}.

\bibitem[Huang et~al.(2015)Huang, Singh, and Ahuja]{urban100}
Jia-Bin Huang, Abhishek Singh, and Narendra Ahuja.
\newblock Single image super-resolution from transformed self-exemplars.
\newblock In \emph{Proc. {IEEE} Conf. Comput. Vis. Pattern Recognit. (CVPR)}, pages 5197--5206, Jun. 2015.
\newblock \doi{10.1109/CVPR.2015.7299156}.

\bibitem[Kingma and Ba(2014)]{adam}
Diederik~P Kingma and Jimmy Ba.
\newblock Adam: A method for stochastic optimization.
\newblock In \emph{Proc. Int. Conf. Learn. Represent. (ICLR)}, May. 2014.

\bibitem[Loshchilov and Hutter(2016)]{cosineannealing}
Ilya Loshchilov and Frank Hutter.
\newblock {SGDR}: Stochastic gradient descent with warm restarts.
\newblock \emph{Proc. Int. Conf. Learn. Represent. (ICLR)}, Apr. 2016.

\bibitem[Wang et~al.(2004)Wang, Bovik, Sheikh, and Simoncelli]{ssim}
Zhou Wang, Alan~C Bovik, Hamid~R Sheikh, and Eero~P Simoncelli.
\newblock Image quality assessment: from error visibility to structural similarity.
\newblock \emph{{IEEE} Trans. Image Process.}, 13\penalty0 (4):\penalty0 600--612, Apr. 2004.
\newblock \doi{10.1109/TIP.2003.819861}.

\bibitem[Wang and Li(2011)]{iw_ssim}
Zhou Wang and Qiang Li.
\newblock Information content weighting for perceptual image quality assessment.
\newblock \emph{{IEEE} Trans. Image Process.}, 20\penalty0 (5):\penalty0 1185--1198, Nov. 2011.

\bibitem[Zhang et~al.(2018{\natexlab{b}})Zhang, Isola, Efros, Shechtman, and Wang]{lpips}
Richard Zhang, Phillip Isola, Alexei~A. Efros, Eli Shechtman, and Oliver Wang.
\newblock The unreasonable effectiveness of deep features as a perceptual metric.
\newblock In \emph{Proc. {IEEE} Conf. Comput. Vis. Pattern Recognit. (CVPR)}, pages 586--595, Jun. 2018{\natexlab{b}}.

\bibitem[Johnson et~al.(2016)Johnson, Alahi, and Fei{-}Fei]{vgg}
Justin Johnson, Alexandre Alahi, and Li~Fei{-}Fei.
\newblock Perceptual losses for real-time style transfer and super-resolution.
\newblock In \emph{Proc. Eur. Conf. Comput. Vis. (ECCV)}, volume 9906, pages 694--711, Oct. 2016.

\bibitem[Zheng et~al.(2022)Zheng, Yuan, Yan, Tian, Zhang, Sun, Liu, Leonardis, and Slabaugh]{freqmoire}
Bolun Zheng, Shanxin Yuan, Chenggang Yan, Xiang Tian, Jiyong Zhang, Yaoqi Sun, Lin Liu, Aleš Leonardis, and Gregory Slabaugh.
\newblock Learning frequency domain priors for image demoireing.
\newblock \emph{{IEEE} Trans. Pattern Anal. Mach. Intell.}, 44\penalty0 (11):\penalty0 7705--7717, Sept. 2022.

\bibitem[Chu et~al.(2022)Chu, Chen, Chen, and Lu]{tlc}
Xiaojie Chu, Liangyu Chen, Chengpeng Chen, and Xin Lu.
\newblock Improving image restoration by revisiting global information aggregation.
\newblock In \emph{Proc. Eur. Conf. Comput. Vis. (ECCV)}, volume 13667, pages 53--71, Oct. 2022.
\newblock \doi{10.1007/978-3-031-20071-7\_4}.

\end{thebibliography}
\end{document}